## RESEARCH ARTICLE

# Signal-to-Noise Ratio in Scanning Electron Microscopy: A Comprehensive Review

**KOK SWEE SIM**[ID]**1, (Senior Member, IEEE), IKSAN BUKHORI**[ID]**2, (Member, IEEE),
DOMINIC CHEE YONG ONG**[ID]**3, (Student Member, IEEE),
AND KOK BENG GAN**[ID]**4, (Member, IEEE)**
[1]Faculty of Engineering and Technology, Multimedia University, Malacca 75450, Malaysia
[2]Department of Electrical Engineering, Faculty of Engineering, President University, Bekasi 17550, Indonesia
[3]School of Engineering, Monash University, Subang Jaya, Selangor 47500, Malaysia
[4]Department of Electrical, Electronic and Systems Engineering, Faculty of Engineering and Built Environment, Universiti Kebangsaan Malaysia,
Bangi 43600, Malaysia
Corresponding author: Kok Swee Sim (kssim@mmu.edu.my)

This work was supported by Multimedia University.

**ABSTRACT** Scanning Electron Microscopy (SEM) is critical in nanotechnology, materials science, and biological imaging due to its high spatial resolution and depth of focus. Signal-to-noise ratio (SNR) is an essential parameter in SEM because it directly impacts the quality and interpretability of the images. SEM is widely used in various scientific disciplines, but its utility can be compromised by noise, which degrades image clarity. This review explores multiple aspects of the SEM imaging process, from the principal operation of SEM, sources of noise in SEM, methods for SNR measurement and estimations, to various aspects that affect the SNR measurement and approaches to enhance SNR, both from a hardware and software standpoint. We review traditional and emerging techniques, focusing on their applications, advantages, and limitations. The paper aims to provide a comprehensive understanding of SNR optimization in SEM for researchers and practitioners and to encourage further research in the field.

## I. INTRODUCTION

Scanning electron microscopy (SEM) is widely used across materials science, biology, and nanotechnology for its ability to reveal three-dimensional surface morphology at sub-micrometres resolution. However, its utility hinges on achieving a high signal-to-noise ratio (SNR), as noise can obscure fine structural details and compromise image interpretation.

As a key instrument in microelectronics research and manufacturing, SEM enables inspection of samples at scales far below one micron. When the primary electron (PE) beam strikes a specimen, it generates multiple types of emissions—secondary electrons (SEs), backscattered electrons (BSEs), Auger electrons (AEs), characteristic X-rays, and photons of various energies ([1], [2])—each originating from different specimen depths and offering distinct structural or compositional insights. In practice, SEM imaging primarily uses SEs and BSEs due to their high surface sensitivity and contrast. Fig. 1 illustrates the fundamental structure of the scanning electron microscope ([3], [4]).

A range of factors—including instrument settings, specimen properties, and detection electronics—can degrade SEM image quality. Noise is particularly detrimental. Although slowing the scan rate (i.e. increasing dwell time per pixel) can reduce noise, it also invites drawbacks such as specimen charging, local contamination, or even damage, and it prolongs settling times.

Quantifying SNR reliably thus requires careful modelling of key noise sources: fluctuations in the primary beam, stochastic generation of SEs or BSEs, and electronic noise from elements like scintillators or photomultiplier tubes. Evaluating SNR in tandem with spatial resolution provides a more complete picture of true image fidelity. To recover hidden detail, modern SEM workflows often combine image-filtering with advanced restoration techniques. This study presents the in-depth review of SEM technology, focusing on its working principles and the studies which have developed over the years to improve the quality of the retrieved images, in particular by reducing the noise presented during image acquisition.

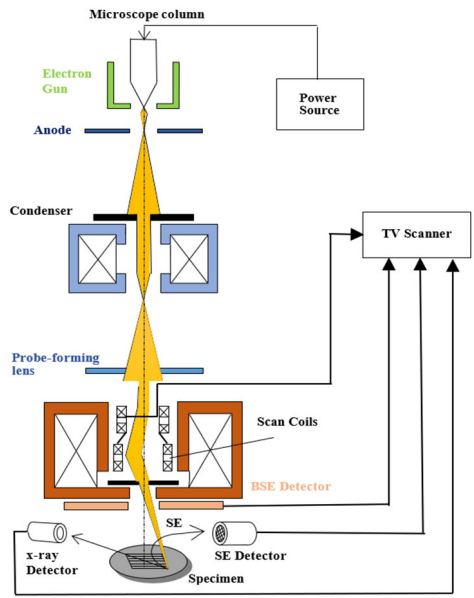

**FIGURE 1.** Working principle of SEM [5].

## II. SNR AND SCANNING ELECTRON MICROSCOPY

This section presents an in-depth exploration of the fundamentals of SNR and Scanning Electron Microscopy.

### A. SIGNAL TO NOISE RATIO (SNR)

The definition of SNR varies across in many different fields. In electrical and electronic engineering. it is defined as the ratio between the energy of the actual signal and the noise signal. For example, in magnetic resonance imaging (MRI), SNR is calculated as the ratio of the mean signal in a region of interest (ROI) to its standard deviation ([6], [7]). In electron microscopy, however, SNR is defined as the ratio between the root mean square (RMS) signal and the RMS fluctuation caused by noise [8].

SNR is used to quantify the noise content in an image. It provides a measure of the signal relative to inherent noise. High-quality images present a high SNR. This parameter is crucial to characterize image quality because noise levels significantly impact image clarity.

Due to the importance of SNR in SEM imaging systems, several SNR estimation techniques have been developed over the years. These techniques are addressed in depth in later sections.

### B. SCANNING ELECTRON MICROSCOPE (SEM)

SEM is a useful tool to produce good resolution images of the surface of a sample. The mechanism in SEM works differently from conventional optical microscopes. SEM applies a focused beam of electrons to obtain high magnification and resolution images, which can provide detailed information about the specimens ([9], [10]).

Various SNR estimation methods and SEM image capture mechanism place a strong emphasis on SNR quantification. Therefore, it is crucial to understand the fundamentals of how SEMs function, the interactions between electron beams and specimens, how images are created by using backscattered and secondary electrons, and the source of noises in SEM images.

### 1) WORKING PRINCIPLE OF SEM

Signal detectors and an electron column form a standard SEM (see Fig. 1). An electron guns, beam-defining apertures, electron lenses, and scanning coils make up the electron column [11]. Between the cathode and anode, the electron gun generates electrons with energy ranging from 1keV to 50keV. For the thermionic guns, the electron beam diameter spans from $10\,\mu m$ to $50\,\mu m$, whereas for field-emission guns, it is between 10 nm and 100 nm [3]. The electron beam in most SEMs interacts with the specimen to produce signals that are used to construct images [12].

A scanning coil system is synchronized with the electron beam of a cathode-ray tube (CRT) to scan the electron probe across a specimen in a raster order through the desired area [13]. The ratio of the CRT's viewing screen size to the area scanned on the specimen can determine the magnification of resulting image. Adjustments to the scan-coil current could achieve the desired magnification while keeping the image size on the CRT consistent.

Today, a scanning coil is connected to a TV scanner rather than being synchronized with a CRT. This setup offers several advantages, including real-time observation on a larger screen and increased accessibility at a lower cost [14]. Through a TV scanner, SEM users can view images immediately on a television display. In addition, TV scanners are affordable and readily available compared to specialized CRT displays. This makes them a cost-effective option for SEM setups [5]. Fig. 1 shows the working principle of SEM and reinforces the described configuration.

### 2) ELECTRON GUNS

The electron gun serves to deliver a stable current within a concentrated electron beam in SEM. It can generate a highly luminous electron source that can be directed onto the surface of a specimen. These electrons interact with the sample, producing either electrons or X-rays that are detectable and used to create SEM images of the sample. SEMs employ various types of electron guns, including thermionic tungsten guns, lanthanum hexaboride (LaB$_6$) guns, Field Emission Electron Guns (FEGs) and Schottky electron guns. Each type differs in its current output, source size, current stability, and longevity.

For many SEM applications where exceptionally high brightness is not necessary but stable high currents are vital, the tungsten filament is a preferred choice due to its cost-effectiveness and reliable performance [15]. Next, LaB₆ sources offer significantly higher brightness and longer filament lifetimes compared to tungsten, but they require a stricter vacuum condition, typically two orders of magnitude better than that needed for tungsten operation [1]

FEGs are chosen when achieving high resolution and optimized low-voltage performance are critical [1]. In these cases, the lifetime of the field emission electron gun is primarily limited by potential damage at the tip caused by electric discharge, assuming the vacuum level remains adequate.

Schottky electron guns blend features from both thermionic and field emission guns. A tungsten filament is used akin to thermionic guns while integrating a unique electrode known as the Schottky emitter, enabling field emission at lower temperatures [16]. This design allows Schottky guns to generate high-brightness electron beams characterized by exceptional stability, making them a prevalent choice in contemporary SEM.

### 3) ELECTRON BEAM-SPECIMEN INTERACTIONS

The electron-specimen interactions in SEM that are relevant for both microanalysis and imaging are illustrated in Fig. 2. It is made up Primary Electrons (PE), Auger Electrons (AE), low loss electrons, Backscattered Electrons (BSE), elastically reflected electrons, and Secondary Electrons (SE). The broad range of BSEs [17] is between 50 eV and primary electron energy ($E = eU$, where $e$ is the electron charge and $U$ is the accelerating voltage).

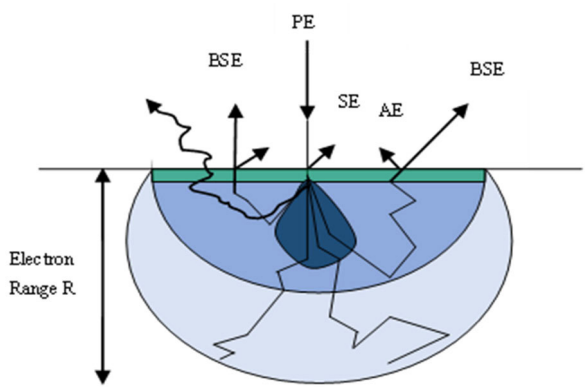

**FIGURE 2.** For the normal incident of primary electrons (PE), information depth of x-ray, backscattered electrons (BSE), secondary electrons (SE) and Auger electrons (AE) in the diffusion cloud of electron range R ([3], [18]).

Secondary emission is defined as the release of electrons from a target material when it is bombarded by an incident PE beam [19]. The emitted electrons include AEs, BSEs, and SEs [3]. Noise from secondary emission arises from the variation in the number of emitted electrons for each incident PE during the interaction with the specimen [20].

Auger Electrons (AEs) have an energy spectrum of 50 eV to 2500 eV being released. Both elastic and inelastic scattering can affect the AEs. With a thickness of a few nanometers of surface layer, these electrons can be reflected from the specimen [3]. If the specimen sample has a band gap, some primary-electron inelastic scattering also excites electrons across it. Subsequent cathodoluminescence (CL) [21] emits visible/UV photons instead of electrons ([22], [23]), offering spectral contrast (e.g. defect lines, dopants), and is collected via optical detectors rather than electron detectors ([22], [23], [24]).

A lot of information is produced by different electron-specimen interactions in terms of quanta and released particles. Different kinds of SEM pictures are formed by all these produced signals. Because SE detectors are frequently used in SEM, most of the researches have focused on SE images [19], [25]. The accuracy and quality of the collected images may be impacted by these noises in SEM images [26]. Consequently, SNR estimation techniques are used to estimate the SNR value of the SEM images to obtain higher-quality images and enhance the accuracy of data based on the photos. Following that, techniques for image filtering are used to remove noise from the SEM pictures.

### C. NOISES IN SEM IMAGE

In SEM, there exist two types of scanning modes namely fast scan mode and slow scan mode. The fast scan mode is to have rapid image acquisition. Usually, it is ideal for swiftly reviewing extensive sample areas or capturing dynamic processes. In fast scan mode, the quick scanning pace compromises spatial resolution and SNR compared to slower scan modes. Subsequently, this trade-off might lead to decrease picture quality and decrease details within the procured pictures [27].

Noise in SEM originates from various types of sources, including the electron beam, detector systems, and environmental factors. A good understanding of these noise types is essential to develop effective SNR enhancement techniques. In some SEM device, a thermionic electron gun is used. In this case, shot noise is the predominant source of noise [28]. This type of noise results from random statistical fluctuations in the number of emitted electrons and is inherent to the primary beam. For SEMs which utilize field emission guns, flicker noise becomes an additional concern. In addition, the detection system, which includes a scintillator and a photomultiplier tube, can also introduce noise. However, this so-called detection noise is relatively insignificant as compared to shot noise and secondary emission noise, as long as the electronic gain remains moderate [28].

Managing noise in SEM images presents a significant challenge. The SNR is influenced by factors such as beam current, the specimen's material composition, and its surface topography. Reimer in [3] explored the emission statistics of secondary and backscattered electrons, while Dubbeldam analyzed shot noise, secondary emission noise, and partition noise in his study [28].

In general, the SEM contains different types of noise, namely shot noise, secondary emission noise, partition noise, emission noise from SE and BSE, thermal noise, quantization noise, and environmental noise.

### 1) SHOT NOISE

Shot noise arises from the quantum nature of electron emission, leading to statistical variations in the number of electrons interacting with the sample. This randomness results in fluctuations in the image signal, particularly at low beam currents. Shot noise follows a Poisson distribution, and its magnitude increases with decreasing electron counts. Shot occurred due to the random arrival of electrons at the detector. The following conditions hold [28] when the electrons emitted directly from an electron source are counted:

1) The distribution of the number of arriving electrons appears to be dependent on only the length of the time interval, but not the initial nor the final instances of the interval.
2) Electrons arrive independently. During any given time interval, the random excess/deficiency of electrons in some time interval does not affect the number of arriving electrons.
3) The probability of more than one electron arriving within a small interval is negligible.

### 2) SECONDARY EMISSION NOISE

Fig. 3 shows the situation where a SE current ($I_{SE}$) is released by a primary-beam current ($I_{PE}$). According to Schottky's theorem [29], the mean square value of noise in the primary electron beam is

$$\bar{i}_{PE}^2 = 2eI_{PE}\Delta f \tag{1}$$

In fact, secondary emission noise is caused by a fluctuation in the number of SEs per primary electron ([30], [31]).

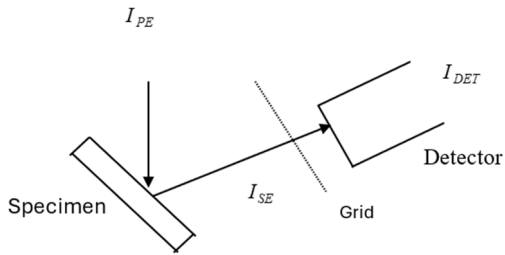

**FIGURE 3.** The origin of secondary emission noise and partition noise [28].

### 3) PARTITION NOISE

In Fig. 2, it shows the Partition noise [28], which is another source of noise that arises from a grid with limited transmissions. If one considers the SEs being divided by a grid with transmission $\gamma$ into a transmitted group and a group is absorbed by the grid, then the noise in the transmitted group is written as Equation (2).

$$\bar{i}_{ddt}^2 = \gamma^2 \bar{i}_{se}^2 + \gamma(1-\gamma)\bar{i}_{se}^2 \tag{2}$$

where $\bar{i}_{se}^2$ is noise in the secondary current and $\gamma$ into a transmitted group.

### 4) EMISSION STATISTICS OF SE AND BSE

In a SEM with a thermionic electron gun, shot noise is the primary source of noise in the primary electron (PE) beam ([32], [33]) [32], [33] and follows Poisson statistics ([3], [34], [35]). For BSE, although the conversion from PE to BSE follows a binomial distribution, the combination of the Poisson statistics of the PE beam and the binomial conversion results in a Poisson distribution for BSE emission.

When the primary electron beam hits a target material, electrons are emitted through secondary emission. These electrons include SEs, BSEs, and AEs. Secondary electrons can further be put into two categories. The first category includes SEs generated by primary electrons entering the specimen, and their secondary emission coefficient is referred to as the SE1 yield, denoted as $\delta_{SE1}$. The second category, the SE2 yield ($\delta_{SE2}$), refers to SEs produced by backscattered electrons as they exit the specimen. Taking into account, the backscattered yield represented by $\varsigma$. The total secondary electron emission coefficient, $\delta$, can be defined as shown in Equation (3) ([36]).

$$\delta = \delta_{SE1} + \delta_{SE2} = \delta_{SE1} + \varsigma\delta_{SE1} \tag{3}$$

In general, shot noise is presented in the PEs. The fluctuations in the number of PEs follow a Poisson distribution. If the number of primary electrons per pixel is denoted as $N_{PE}(x, y)$, then

$$N_{PE}(x, y) = \bar{N}_{PE}(x, y) + f(N_{PE}) \tag{4}$$

where $\bar{N}_{PE}(x, y)$ is the mean number of primary electrons per pixel and $f(N_{PE})$ is the fluctuation in the number of primary electrons per pixel.

In the absence of noise, the number of secondary electrons (SEs) emitted per pixel is $N_{SE}^{NF}(x, y)$.

$$N_{SE}^{NF}(x, y) = \delta.(\bar{N}_{PE}(x, y) + f(N_{PE})) = \delta.(N_{PE}(x, y)) \tag{5}$$

where $\delta$ is the SE yield. The SE electrons are excited by PEs with a yield of $\delta$ is as shown in Equation (6). Similarly, the number of noise free BSEs electrons per pixel, $N_{BSE}^{NF}(x, y)$, can be Equation (7) [36]

$$N_{BSE}^{NF}(x, y) = \varsigma.(\bar{N}_{PE}(x, y) + f(N_{PE})) = \eta.(N_{PE}(x, y)) \tag{6}$$

where $\varsigma$ is the BSE emission yield

If there is noise in the secondary electron emission, the number of SEs electrons per pixel is $N_{SE}(x, y)$ and it is shown in Equation (7) [36]

$$N_{SE}(x, y) = (\delta + RV(\delta))(\bar{N}_{PE}(x, y) + f(N_{PE})$$
$$= (\delta + RV(\delta))(N_{PE}(x, y)) \tag{7}$$

where $RV(\delta)$ is a random variable. It represents the instantaneous SE yield with mean and variance follow a Poisson Distribution.

The number of BSEs per pixel with emission noise, $N_{BSE}(x, y)$, is shown at Equation (8)

$$N_{BSE}(x, y) = (\varsigma + RV(\eta))(\bar{N}_{PE}(x, y) + f(N_{PE}))$$
$$= (\varsigma + RV(\eta))(N_{PE}(x, y)) \tag{8}$$

where $RV(\varsigma)$ is the random variable. It represents the BSE yield with mean and variance values follow a Binomial Distribution.

### 5) THERMAL NOISE
Thermal noise is a consequence of random thermal agitation of charge carriers in electronic components, such as detectors and amplifiers. This noise type is characterized by a Gaussian distribution and contributes to image degradation in SEM systems, especially at low signal levels.

### 6) QUANTIZATION NOISE
Quantization noise occurs during the digitization process when the analog signal from the detector is converted into discrete pixel values. The rounding errors inherent in this conversion introduce noise, which becomes more pronounced at lower resolution levels.

### 7) ENVIRONMENTAL NOISE
Environmental noise includes external factors such as mechanical vibrations, acoustic interference, and electromagnetic fields. These can introduce artifacts such as image drift into the SEM image, especially in high-magnification settings.

## III. DIRECT SNR ESTIMATION IN SEM
In this section, we discuss related SEM works on the estimation of SNR in SEM images. Accurate estimation of SNR is critical for assessing image quality and the effectiveness of noise reduction techniques.

Despite advances in SEM technology, several challenges persist in maintaining an optimal SNR. These include fluctuations in electron beam currents, the complex interaction between the electron beam and various sample materials, and environmental factors such as electromagnetic interference and vacuum quality. Traditional methods for improving SNR often involve trade-offs, such as increasing beam intensity at the cost of sample damaged or utilizing difference scan rate to estimate signal and noise, which can be time-consuming and computationally expensive.

However, traditional SNR enhancement techniques do not fully address the dynamic and variable nature of noise in different imaging conditions. The lack of a unified framework for accurately estimating and improving SNR across various SEM applications results in inconsistent image quality, impeding precise scientific measurements and analyses. Therefore, a comprehensive review of current methodologies is important to identify the limitations of existing techniques, propose novel solutions, and guide future advancements in SEM technology.

When the noise is actually known or can safely be assumed (e.g., the gaussian noise with certain variance and mean), one can compare the signal and the noise directly. There are two different common methods in calculating SNR in this way.

### A. RATIO OF VARIANCES
One common method is to compute the ratio of the variance of the image signal to the variance of the noise. While straightforward, this method assumes that the noise is uniformly distributed across the image, which may not always be the case in SEM images due to spatial variations in signal intensity.

### B. FOURIER TRANSFORM-BASED METHODS
Fourier-based techniques can separate high-frequency noise from the low-frequency signal. By applying a Fourier transform to the SEM image, the signal and noise components can be analyzed in the frequency domain, allowing for the calculation of SNR based on the power spectra. These methods are effective but require prior knowledge of the noise characteristics.

However, in many cases such assumption or knowledge about the noise may not be present. To solve this problem, various methods have been developed for SNR estimation in SEM images, some of which are rooted in the general parameter estimation techniques ([37], [38]).

Quantifying the SNR is crucial in image acquisition processes, especially in electron microscopy and other fields where images are affected by noise. In SEM, there is often a trade-off between image resolution and SNR.

The cross-correlation [39] is one of the earliest proposed techniques for SNR estimation in microscopic images. Researchers ([6], [40], [41], [42]) applied the same technique of Frank and Al-Ali in microscopic and MRI images.

In 2004, Sim developed a single image SNR estimation technique to determine the SNR of microscopic images. In another approach, [3] and [43] proposed to measure the SEM images by measuring the Secondary Electron (SE) yield.

Among all the above techniques, they can be classified as two image SNR measurement technique, a single image SNR estimation technique, and the SE yield SNR measurement.

Techniques have been developed through the years to measure the Image SNR in Scanning Electron Microscopy. Table 1 lists various classes of method available to measure and estimate SNR.

For two images SNR measurement approaches, there are Frank and Al-Ali Method [39], and Scanning Microscope Analysis and Resolution Testing (SMART) ([39], [41]). For single image SNR estimation approaches, there are Simple method [44], First order interpolation [44], Linear Least Squares Regression (LSR) [45]; Non-linear Least Squares Regression (NLLSR) [46], Adaptive Slope Nearest Neighbourhood (ASNN) [47]; Autocorrelation Levinson–durbin Recursion (ACLDR) [48], Cubic Hermite Interpolation with Linear Least Square Regression Signal-to-noise ratio (CHILLSRSNR) ([49], [50]). In term of hardware measurement, SE yield SNR measurement is another approach [33].

The following sections describe the various techniques, the application classes of SNR measurement and estimation as well as the advantages and fundamental limitations of each method.

**TABLE 1.** Various SNR estimation and measurement techniques.

| Category | Methods | | | | | | | Remarks |
|---|---|---|---|---|---|---|---|---|
| Direct Estimation | Ratio of Variances | Fourier Transform-based Estimation | | | | | | Can be used when noise is known or can be safely assumed |
| Two Images Estimation | Frank and Al-Ali Method [39] | SMART ([39][41]) | | | | | | Can be used when two images are available, one of which can be thought of noiseless image |
| Single Image Estimation | Simple method [53] | FOL [54] | LSR [45] | NLLSR [46] | ASNN [47] | ACLDR [48] | CHILLSRSNR ([49], [50]) | Can be used when there is only a single image and the noise is unknown |
| Hardware-based Measurement | SE yield SNR measurement [33] | | | | | | | Can be used when the hardware can be set up in a precise way |

## IV. TWO IMAGES SNR ESTIMATION METHODS
### A. FRANK AND AL-ALI METHOD
In 1975, Frank [39] developed the theoretical basis for the two-image SNR estimation. The idea was to perform two image acquisitions of the same object. Then, the cross-correlation coefficient between these two images can be computed using Equation (9).

$$\rho_{12} = \frac{r_{12}(0,0) - \mu_1 \mu_2}{\sigma_1 \sigma_2} \qquad (9)$$

Here, $r_{12}(0,0)$ represents the peak of the cross-correlation function (CCF) of the two aligned images. $\mu_1$ and $\mu_2$ are the means for the first and the second image, respectively. $\sigma_1$, $\sigma_2$ are the variances of the corresponding images. The SNR is then given as

$$SNR = \frac{\rho_{12}}{1 - \rho_{12}} \qquad (10)$$

One of the underlying assumptions of this method is that both image acquisitions contain the same signal with uncorrelated noise of zero mean. The process of estimating the SNR involves selecting two images from the experimental image set, align them and evaluate them using Equation (10). To minimize variance in the SNR estimate, an average of estimates can be obtained by randomly selecting pairs of images [51].

For a while, determining SNR using two image acquisitions to is a widely used method. However, the vast application of the cross-correlation function in SNR estimation reveals two limitations in such technique: (1) perfect alignment between the two images is required, and (2) these methods cannot be used to determine the SNR of an existing image, such as a stored image or micrograph.

### B. SCANNING MICROSCOPE ANALYSIS AND RESOLUTION TESTING (SMART)
In 2000, Joy et al. [42] utilized CCF to evaluate the SNR performance of SEM. The use of CCF ([39], [41]) provides a useful alternative approach to the Fourier analysis as it can avoid the problem to distinguish signal from noise. The CCF itself can be written as

$$c(x,y) = f(x,y) \otimes g(x,y) \qquad (11)$$
$$C(u,v) = F(u,v)G^{\wedge} * (u,v) \qquad (12)$$

In this context, $F(u,v)$ and $G(u,v)$ represent the two-dimensional power spectra of the images $f(x,y)$ and $g(x,y)$ respectively, while $C(u,v)$ denotes the product of $F(u,v)$ and the conjugate of $G(u,v)$. When $f(x,y)$ and $g(x,y)$ are two samples of the same image which are separated by a few pixels, the Cross-Correlation Function (CCF) exhibits a sharp peak. The displacement of this peak from the center reflects the pixel offset between the images, with its full width at half maximum (FWHM) corresponding to the Rayleigh criterion for image resolution.

This shows that image details are correlated over a distance equal to the image resolution, whereas noise, being random, remains uncorrelated from pixel to pixel. The peak-to-background ratio of the CCF can be used to estimate the SNR of the image itself.

The Scanning Microscope Analysis and Resolution Testing (SMART) program was developed by [42]. The process begins by selecting a region of interest within the image. The program computes the Fast Fourier Transform (FFT) of this region and then selects a second region of the same size, shifted by a few pixels. The FFT of the second region is also calculated. By using Equation (12), the Cross-Correlation Function (CCF) is then computed. A line profile is drawn through the CCF peak. The image resolution and SNR are determined by analyzing the peak measurements.

In Fig. 4, it illustrates the application of this method to the IC sample material. Fig. 4(a) shows the selected region of interest (ROI) from the IC image, while Fig. 4(b) depicts the computed Cross-Correlation Function (CCF). The CCF is represented as a two-dimensional image, where brightness corresponds to the value of $c(j, k)$ as defined in Equation (13). Fig. 5 shows the line profile across the CCF peak.

The SMART method employs the cross-correlation function to assess the SNR performance of the SEM. However, there is a significant limitation of this method as the requirement for perfect alignment of the two images is needed.

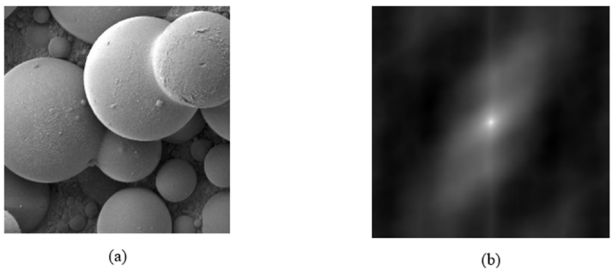

**FIGURE 4.** Cross-correlation Function analysis of SEM image at (a) the analyzed Region of Interest; (b) the cross-correlation function plot. Horizontal field-width = 50 $\mu m$ and beam energy = 10 keV. Image size is 512 by 512 pixels.

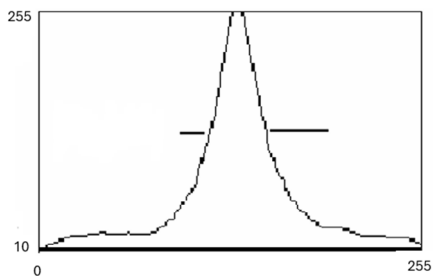

**FIGURE 5.** The intensity plot across the CCF peak shows the definition of the resolution as the peak full width half maximum.

## V. SINGLE IMAGE SNR ESTIMATION TECHNIQUES

The method to derive the SNR of a single image starts with Equation (9) and (10) for two images of identical signal but uncorrelated noise was implemented by [36] and [44]. It starts with Equation (9), where $r_{12}(0, 0)$ is the CCF between images $g_1(x, y) = s_1(x, y) + n_1(x, y)$ and $g_2(x, y) = s_2(x, y) + n_2(x, y)$ at zero offset, where $s_1$ and $s_2$ are the noise-free images, and $n_1$ and $n_2$ represent the noise content of these two images.

Since $n_1$ and $n_2$ are uncorrelated, and the noise is assumed to be uncorrelated with the signal, we have

Equation (13)

$$
\begin{aligned}
r_{12}(0, 0) &= \frac{1}{N^2} \sum_{j=0}^{N-1} \sum_{i=0}^{N-1} (s_1(i, j) \\
&\quad + n_1(i, j)) (s_2(i, j) + n_2(i, j)) \\
&= \frac{1}{N^2} \sum_{j=0}^{N-1} \sum_{i=0}^{N-1} s_1(i, j) \cdot s_2(i, j) \\
&= \bar{r}_{12}(0, 0)
\end{aligned}
\tag{13}
$$

where $\bar{r}_{12}(0, 0)$ denotes the CCF between the two noise-free images, $s_1$ and $s_2$, at zero offset. Furthermore, since the noise-free images are identical, $s_1 = s_2$ and

$$
r_{12}(0, 0) = \bar{r}_{12}(0, 0) = \bar{r}_{11}(0, 0)
\tag{14}
$$

where $\bar{r}_{11}(0, 0)$ is the value of the autocorrelation function (ACF) of the noise-free image at zero offset. The ACF is defined when $s_1$ and $s_2$ are the two identical functions.

It should be noted that the mean values of both images are identical, since both images have identical signal corrupted with noise of zero mean. Therefore,

$$
\mu_1 = \mu_2
\tag{15}
$$

The variance of image 1, $\sigma_1$, is given as

$$
\sigma_1^2 = \frac{1}{N^2} \sum_{j=0}^{N-1} \sum_{i=0}^{N-1} (f_1(i, j) - \mu_1)^2 = r_{11}(0, 0) - \mu_1^2
\tag{16}
$$

Similarly, for image 2,

$$
\sigma_2^2 = r_{22}(0, 0) - \mu_2^2
\tag{17}
$$

Since $r_{11}(0, 0) = r_{22}(0, 0)$, it is obtained that

$$
\sigma_1^2 = \sigma_2^2
\tag{18}
$$

Taking square root on both sides of Equation (20), we have

$$
\sigma_1 = \sigma_2
\tag{19}
$$

From Equations (16)-(21)

$$
\rho_{12} = \bar{\rho}_{11} = \frac{\bar{r}_{11}(0, 0) - \mu_1^2}{\sigma_1^2}
\tag{20}
$$

And

$$
\begin{aligned}
SNR &= \frac{\rho_{12}}{1 - \rho_{12}} = \frac{\bar{r}_{11}(0, 0) - \mu_1^2}{\sigma_1^2 - \bar{r}_{11}(0, 0) + \mu_1^2} \\
&= \frac{\bar{r}_{11}(0, 0) - \mu_1^2}{r_{11}(0, 0) - \bar{r}_{11}(0, 0)}
\end{aligned}
\tag{21}
$$

Thus, the SNR of a single image can be obtained from the ACF curve as shown in Fig. 6.

As shown in Fig. 6, $r_{11}(0, 0) - \bar{r}_{11}(0, 0)$ is the noise energy and $\bar{r}_{11}(0, 0) - \mu_1^2$ is the signal energy, where $\mu_1$ is the mean value of image,

$$
\mu_1 = \frac{1}{N^2} \sum_{j=0}^{N-1} \sum_{i=0}^{N-1} f_1(i, j)
\tag{22}
$$

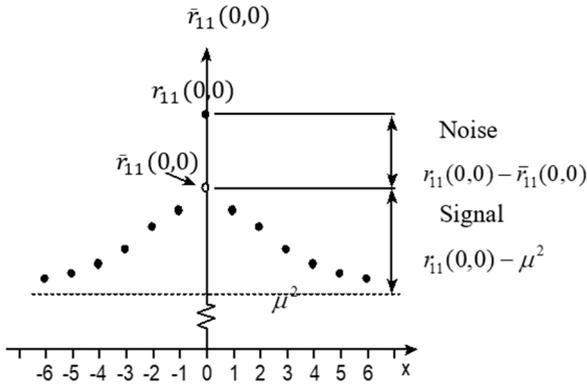

**FIGURE 6.** Representation of signal and noise components on a plot of the autocorrelation function. The filled markers represent the data derived from the image [36].

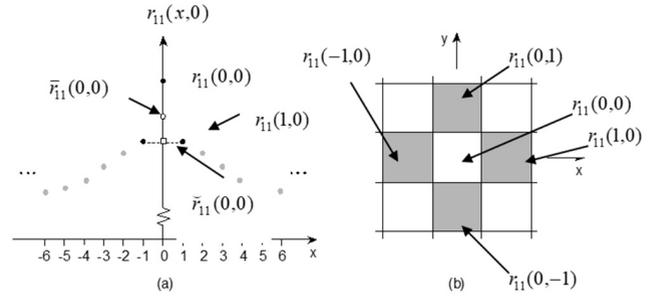

**FIGURE 7.** Estimation of $\bar{r}_{11}(0, 0)$ by assigning the autocorrelation function values at neighboring offsets (a) along the x-direction, (b) along both x and y direction [44].

Unfortunately, with a single image corrupted by noise, $(s_1 + n_1)$, $\bar{r}_{11}(0, 0)$ cannot be obtained directly since the noise $n_1$ is correlated at zero offset and

$$
\begin{aligned}
r_{11}(0, 0) &= \frac{1}{N^2} \sum_{j=0}^{N-1} \sum_{i=0}^{N-1} (s_1(i, j) \\
&\quad + n_1(i, j))(s_1(i, j) + n_1(i, j)) \\
&= \frac{1}{N^2} \sum_{j=0}^{N-1} \sum_{i=0}^{N-1} s_1^2(i, j) + n_1^2(i, j) \\
&\neq \bar{r}_{11}(0, 0)
\end{aligned}
\tag{23}
$$

Since $\bar{r}_{11}(0, 0)$ is unknown, a method for estimating $\bar{r}_{11}(0, 0)$ is required. Several methods had been developed to predict the noise-free zero offset point [52]. For simplicity, we can show the correlation function with offset along the x direction at zero y offset. So, the two-dimensional autocorrelation function $r(x, y)$ may be reduced into a single dimensional function, $r(x, 0)$.

### A. SIMPLE METHOD

In the first method, Sim et al. [53] estimated the power of noise free image, $\bar{r}(0, 0)$, denoted as $\check{r}(0, 0)$, equating it to one of the two adjacent autocorrelation function (ACF) at neighboring offsets (Fig. 7a), namely, $r(1, 0)$ or $r(-1, 0)$. Thus, we have Equation (24)

$$
\check{r}(0, 0) \approx r(1, 0) = r(-1, 0)
\tag{24}
$$

Consider a two-dimensional case as shown in Fig. 7b. Considering the unit offset in the x and y directions of ACF, the values can be averaged as shown in Equation (25)

$$
\check{r}(0, 0) \approx \frac{r(1, 0) + r(0, 1)}{2}
\tag{25}
$$

This estimation shows reasonable results only if the autocorrelation function of the noisy image changes slowly around the origin, which is applicable for images where the details are correlated over many pixels.

### B. FIRST ORDER INTERPOLATION (FOL) METHOD

An alternative approach involves using first-order interpolation [54]. In the $x-$ direction, as illustrated in Fig. 8, the adjacent points $r_{11}(1, 0)$ and $r_{11}(2, 0)$ can be utilized to predict $\bar{r}_{11}(0, 0)$. This approach provides a better estimate compared to Equation (22). Higher-order functions, such as polynomials, can be employed to fit the Auto-Correlation Function (ACF) curve. However, while both methods are viable, they tend to have limited accuracy and are dependent on the characteristics of the images.

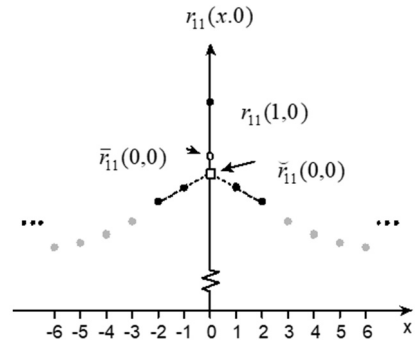

**FIGURE 8.** Estimation of $\bar{r}_{11}(0, 0)$ by a first-order interpolation from neighboring offsets along x-direction [44].

### C. LINEAR LEAST SQUARE REGRESSION (LSR) METHOD

The Linear Least Squares Regression (LSR) method was proposed by Sim and Norhisham in [45] to estimate the SNR. As shown by the autocorrelation function (ACF) curve in Fig. 8, it is evident that the value of the point preceding the peak is always less than the peak itself. After computing the autocorrelation function, the estimated SNR value consistently falls within the confidence interval, as illustrated in Fig. 9.

The equation is derived from the straight-line equation, as shown in Equation (26), where $\alpha$ denotes the y-intercept and $\beta$ represents the slope [45].

$$
\check{\gamma} = \alpha + \beta X
\tag{26}
$$

The value of $\hat{y}$ is calculated for SNR estimation using this method. $\check{\gamma}$ is assumed to represent the predicted noise-free peak. To account for the random error in this method, $\varepsilon$ is introduced into Equation (26) to represent the unpredicted error, resulting in the formation of Equation (27).

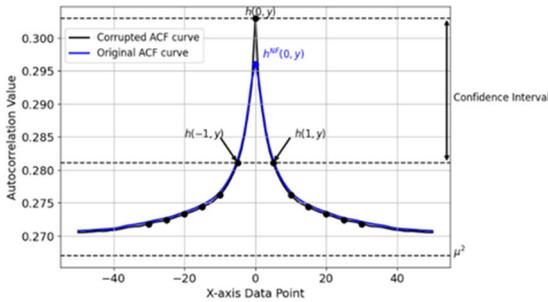

**FIGURE 9.** Confidence interval of estimated SNR value [55].

$$\check{\gamma} = \alpha + \beta X + \varepsilon \qquad (27)$$

In Matrix form, Equation (27) can be written as

$$\check{\gamma} = XB + \boldsymbol{\varepsilon} \qquad (28)$$

where

$$\check{\gamma} = \begin{bmatrix} r_1 \\ r_2 \\ \vdots \\ r_N \end{bmatrix}, \boldsymbol{B} = \begin{bmatrix} \alpha \\ \beta \end{bmatrix}, X = \begin{bmatrix} 1 & x_1 \\ 1 & x_2 \\ . & . \\ . & . \\ 1 & x_N \end{bmatrix} \text{ and } \boldsymbol{\varepsilon} = \begin{bmatrix} \varepsilon_1 \\ \varepsilon_2 \\ \vdots \\ \varepsilon_N \end{bmatrix}$$

Such that

$$\begin{bmatrix} r_1 \\ r_2 \\ . \\ . \\ r_N \end{bmatrix} = \begin{bmatrix} \alpha x_1 + \beta + \varepsilon_1 \\ \alpha x_2 + \beta + \varepsilon_2 \\ . \\ . \\ \alpha x_N + \beta + \varepsilon_N \end{bmatrix} \qquad (29)$$

To reduce the unpredicted error, $\boldsymbol{\varepsilon}$ in Equation (29), we need to have

$$\left[\min_\beta \sum_{k=1}^N \varepsilon_k^2 = \min_\beta \boldsymbol{\varepsilon}^T \boldsymbol{\varepsilon}\right] = \left[\frac{d}{d\beta} \sum_{k=1}^N \varepsilon_k^2 = \frac{d}{d\beta} \boldsymbol{\varepsilon}^T \boldsymbol{\varepsilon}\right] = 0 \qquad (30)$$

Let $\boldsymbol{\varepsilon} = \check{\gamma} - XB$. The minimization problem can be solved as

$$\frac{d}{d\beta}\left(\boldsymbol{\varepsilon}^T \boldsymbol{\varepsilon}\right) = \frac{d}{d\beta}\left(\check{\gamma} - XB\right)^T \left(\check{\gamma} - XB\right) = 0$$

$$\frac{d}{d\beta}\left(\boldsymbol{\varepsilon}^T \boldsymbol{\varepsilon}\right) = \frac{d}{d\beta}\left(\check{\gamma} - XB\right)^T \left(\check{\gamma} - XB\right) = 0$$

$$\frac{d}{d\beta}\left(\check{\gamma}^T \check{\gamma} - \check{\gamma}^T XB + \boldsymbol{B}^T X^T XB - \check{\gamma}^T XB\right) = 0$$

$$2X^T XB - 2X^T \check{\gamma} = 0$$

$$X^T XB = X^T \check{\gamma}$$

$$\boldsymbol{B} = (X^T X)^{-1} X^T \check{\gamma} = \begin{bmatrix} \alpha \\ \beta \end{bmatrix} \qquad (31)$$

The vector Equation (31) can then be transformed into matrix form, yielding the Equation (32).

$$\begin{bmatrix} \propto \\ \beta \end{bmatrix} = \left[ \begin{pmatrix} x_1 & x_2 & \dots & x_N \\ 1 & 1 & \dots & 1 \end{pmatrix} \begin{pmatrix} x_1 & 1 \\ x_2 & 1 \\ \vdots & \vdots \\ x_N & 1 \end{pmatrix} \right]^{-1}$$

$$\begin{pmatrix} x_1 & x_2 & \dots & x_N \\ 1 & 1 & \dots & 1 \end{pmatrix} \begin{pmatrix} y_1 \\ y_2 \\ \vdots \\ y_N \end{pmatrix} \qquad (32)$$

The values of $\alpha$ and $\beta$ are determined by selecting points from the y-axis, while points from the x-axis of the ACF curve are substituted into Equation (32). In the ACF curve, the x-axis points always increase linearly, as illustrated in Fig. 10. When four points from the x-axis and y-axis are chosen, N = 4, and the points must be selected in a linear sequence [46], as depicted in Fig. 11.

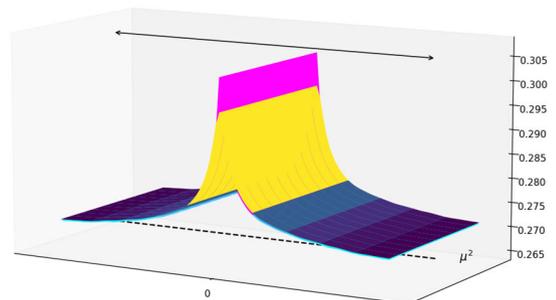

**FIGURE 10.** Points of x-axis of the ACF always increase linearly in the ACR curve [55].

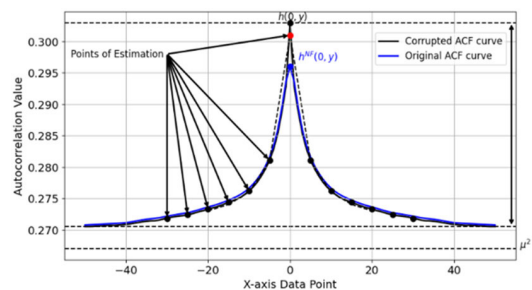

**FIGURE 11.** Points of y-axis randomly increase in the ACF curve [55].

While Equation (33) displays the equation of $\varepsilon$, Equation (34) is the noise-free peak equation. More research is needed to determine the precise value of $\varepsilon$ because it varies depending on the type of SEM images [55].

$$\varepsilon = \frac{h(0, y) - h(1, y)}{2} \qquad (33)$$

$$h^{LSR}(0, y) = \hat{Y} = \alpha + BX + \varepsilon \qquad (34)$$

Equation (35) is the final equation based on LSR method for estimating the SNR value of SEM images.

$$SNR = \frac{h^{LSR}(0, y) - u^2}{h(0, y) - h^{LSR}(0, y)} = \frac{(\alpha + BX + \varepsilon) - u^2}{h(0, y) - (\alpha + BX + \varepsilon)} \qquad (35)$$

The estimated SNR results show that instead of overestimating or underestimating the actual SNR values, the LSR technique tends to follow the shape of the ACF curve. The noiseless peak is determined by the LSR technique using the points that precede or follow the noisy peak. As a result, the number of points (N) must be chosen carefully in order to have a satisfactory SNR estimation accuracy. Prior to the noiseless peak, Sim and Norhisham calculated the SNR value up to five points.

### D. NON-LINEAR LEAST SQUARES REGRESSION (NLLSR) METHOD

Another approach that Sim & Norhisham proposed in 2016 [46] is NLLSR. As seen in Fig. 11, the ACF curve exhibits a slight exponential increasing relationship. As illustrated in Fig. 12, the rationale states that the first and second quadrants exhibit modest exponential growth [56].

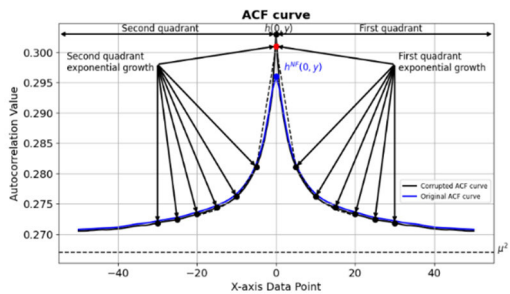

**FIGURE 12.** Slight exponential growth on first and second quadrant ([46], [55]).

Equation (36) is formed by transforming Equation (27) into a non-linear form.

$$ln\ \check{\gamma} = ln(\alpha) + \beta X + ln(\varepsilon) \qquad (36)$$

Here $\alpha$ is the initial beginning amount, $\beta$ is the relative growth rate, $X$ is the x-axis of the ACF curve, $\hat{\gamma}$ is the anticipated noiseless peak, $h_{NF}(0, y)$, and $\varepsilon$ is the random white noise, which together define the equation of continuous growth or decay [46]. Therefore, $\varepsilon$ will not be zero or negative in a distorted SEM image. Thus, the solution to Equation (36) can be written as

$$\check{\gamma} = (\alpha)\ (\varepsilon)\ e^{\beta X} \qquad (37)$$

Only when the ACF curve exhibits a continuous increase or decline analogy is Equation (38) relevant. It is reasonable to assume that each point will rise as it approaches the peak in the second quadrant of the ACF curve and fall as it moves away from $h(0, y)$ in the first quadrant [46]. The growth or decay requirement is satisfied by this result; however, the random error value, represented by $\varepsilon$, which is influenced by the beam diameter and accelerating voltage, determines the degree of stability in growth or decay [49]. The estimated noiseless peak in a corrupted or noiseless SEM picture can never be zero or negative, unlike the x-axis value, which can be positive, negative, or zero. As a result, in this case, x-axis numbers are deemed to be positive. This allowed for the application of Equation (38) [46], often known as the constant elasticity equation. By applying a logarithm to the x-axis, it is now possible to follow the non-linear trend. This is due to the lack of a definition for the natural logarithm of values that are zero or negative on the x-axis.

$$ln\ \check{\gamma} = ln\ (\alpha) + (\beta)ln\ (X) + ln\ (\varepsilon) \qquad (38)$$

In order to create the SNR estimation formula for the NLLSR approach, Equation (32) has been simplified to be as shown in Equation (9). The number of points chosen here is denoted by $N$.

$$\begin{bmatrix} \propto \\ \beta \end{bmatrix} = \left[ \begin{pmatrix} X_1 & X_2 & \dots & X_N \\ 1 & 1 & \dots & 1 \end{pmatrix} \begin{pmatrix} X_1 & 1 \\ X_2 & 1 \\ \vdots & \vdots \\ X_N & 1 \end{pmatrix} \right]^{-1}$$

$$\begin{pmatrix} X_1 & X_2 & \dots & X_N \\ 1 & 1 & \dots & 1 \end{pmatrix} \begin{pmatrix} Y_1 \\ Y_2 \\ \vdots \\ Y_N \end{pmatrix}$$

$$= \begin{bmatrix} N & \sum\limits_{i=1}^{N} X_i \\ \sum\limits_{i=1}^{N} X_i & \sum\limits_{i=1}^{N} X_i^2 \end{bmatrix}^{-1} \begin{pmatrix} \sum\limits_{i=1}^{N} Y_i \\ \sum\limits_{i=1}^{N} Y_i X_i \end{pmatrix} \qquad (39)$$

The first order NLLSR technique Equation (40) is then produced by modifying Equation (39) by the logarithm property [46].

$$\begin{bmatrix} \propto \\ \beta \end{bmatrix}$$
$$= \begin{bmatrix} N & \sum_{i=1}^{N} ln\ (X_i) \\ \sum_{i=1}^{N} ln\ (X_i) & \sum_{i=1}^{N} ln\ (X_i)^2 \end{bmatrix}^{-1} \begin{pmatrix} \sum_{i=1}^{N} ln\ (Y_i) \\ \sum_{i=1}^{N} ln\ (Y_i)\ ln\ (X_i) \end{pmatrix} \qquad (40)$$

The higher order NLLSR can then described in as in Equation (41), as shown at the top of the next page.

In this equation, $M$ represents the order. This number must be carefully determined to avoid matrix singularity. Non-linear functions are far more effective in avoiding matrix singularity than linear functions [46]. Equation (42) illustrates how the higher order $\check{\gamma}$, which corresponds to $h_{NF}(0, y)$, is produced based on Equation (43).

$$ln\ \check{\gamma} = ln\ (\alpha) + (\beta_1)\ ln\ (X)$$
$$+ (\beta_2)\ ln\ \left(X^2\right)\dots$$
$$+ (\beta_M)\ ln\ \left(X^M\right)$$
$$+ ln\ (\varepsilon) \qquad (42)$$

Syafiq in [46], however, only employs the first order NLLSR method for SNR estimation.

$$\check{Y} = \alpha\varepsilon \prod_{k=1}^{M} \left(X^k\right)^{\beta_k} \qquad (43)$$

$$\begin{bmatrix} \propto \\ \beta_1 \\ \beta_2 \\ \vdots \\ \beta_M \end{bmatrix} = \begin{bmatrix} N & \sum_{i=1}^{N} \ln(X_i)^k & \sum_{i=1}^{N} \ln(X_i)^{k+1} & \cdots & \sum_{i=1}^{N} \ln(X_i)^M \\ \sum_{i=1}^{N} \ln(X_i)^k & \sum_{i=1}^{N} \ln(X_i)^{k+1} & \sum_{i=1}^{N} \ln(X_i)^{k+2} & \cdots & \sum_{i=1}^{N} \ln(X_i)^{M+1} \\ \sum_{i=1}^{N} \ln(X_i)^{k+1} & \sum_{i=1}^{N} \ln(X_i)^{k+2} & \sum_{i=1}^{N} \ln(X_i)^{k+3} & \cdots & \sum_{i=1}^{N} \ln(X_i)^{M+2} \\ \vdots & \vdots & \vdots & \ddots & \vdots \\ \sum_{i=1}^{N} \ln(X_i)^M & \sum_{i=1}^{N} \ln(X_i)^{M+1} & \sum_{i=1}^{N} \ln(X_i)^{M+2} & \cdots & \sum_{i=1}^{N} \ln(X_i)^{2M} \end{bmatrix}^{-1} \begin{pmatrix} \sum_{i=1}^{N} \ln(Y_i) \\ \sum_{i=1}^{N} \ln(Y_i) \ln(X_i)^k \\ \sum_{i=1}^{N} \ln(Y_i) \ln(X_i)^{k+1} \\ \vdots \\ \sum_{i=1}^{N} \ln(Y_i) \ln(X_i)^M \end{pmatrix} \tag{41}$$

According to Fig. 8, the noise-free peak at $X = 0$ is represented by $h_{NF}(0, y)$. However, Equation (44) will yield a zero or merely an $\alpha$, respectively, when $h_{NF}(0, y)$ is at $X = 0$ or $X = 1$. Therefore, a distinct set of points will be chosen [46].

The value of the $\alpha$ and $\beta$ coefficients is then obtained by substituting the selected points into Equation (40) after that. Equation (43) is then obtained in order to estimate $\hat{Y}$ Then, Equation (44) is represented as $\hat{Y}^{NLLSR}$.

$$\check{\gamma}^{NLLSR} = \check{\gamma} = \alpha \varepsilon \prod_{k=1}^{M} \left( X^k \right)^{\beta_k} \tag{44}$$

Lastly, by replacing Equation (44) into Equation (17), we have

$$SNR = \frac{\check{\gamma}^{NLLSR} - u^2}{h(0, y) - \check{\gamma}^{NLLSR}} = \frac{\left[ \alpha \varepsilon \prod_{k=1}^{M} \left( X^k \right)^{\beta_k} \right] - u^2}{h(0, r) - \left[ \alpha \varepsilon \prod_{k=1}^{M} \left( X^k \right)^{\beta_k} \right]} \tag{45}$$

### E. ADAPTIVE SLOPE NEAREST NEIGHBOURHOOD (ASNN) METHOD

An SNR estimate technique known as the ASNN approach was proposed by Sim and Teh [47] in 2015. It made use of the NN method with straight line equation. Equation (46) is applied for this strategy (based on Equation (18)).

$$SNR_{actual} = \frac{r^{NF}(0, y) - \mu^2}{r(0, y) - r^{NF}(0, y)} \tag{46}$$

The straight-line equation ($r = SX + c$) was used to create Equation

$$SNR_{predicted} = (S)SNR_{actual} - c \tag{47}$$

The variable $S$ is defined as

$$S = \frac{r^{NF}(0, y)}{r(0, y)} \tag{48}$$

Equation (49) shows the ASNN method's general equation

$$SNR_{predicted} = (0.99744)SNR_{actual} - 0.00645 \tag{49}$$

The NN, FOL, and SP2CHARMA methods were contrasted with the ASNN approach in [49]. In SNR estimation, the ASNN technique performs better than the NN, FOL, and SP2CHARMA methods since it is not impacted by the image's characteristics. In contrast to the original SNR values, it also provides excellent accuracy and a low percentage of estimation error, regardless of the noise variance values [57].

### F. AUTOCORRELATION LEVINSON–DURBIN RECURSION (ACLDR) METHOD

Sim et al. proposed the ACLDR approach [48] in 2016. The predicted noise-free peak for the ACLDR technique is obtained by fitting the Levinson order-update equation to the autocorrelation value [48].

We start from Toeplitz equation and derive

$$a_{n+1} v_{n+1} = \varepsilon_{n+1} \tag{50}$$

Which can be expanded into Equation (51)

$$\begin{bmatrix} a_q(0) & a_q*(1) & \cdots & a_q*(n+1) \\ a_q(1) & a_q(0) & \cdots & a_q*(n) \\ \vdots & \vdots & \cdots & \vdots \\ a_q(n) & a_q(n-1) & \ddots & a_q*(1) \\ a_q(n+1) & a_q(n) & \cdots & a_q(0) \end{bmatrix} \begin{bmatrix} 1 \\ v_n(1) \\ \vdots \\ v_n(n) \\ 0 \end{bmatrix} = \begin{bmatrix} \varepsilon_n \\ 0 \\ \vdots \\ 0 \\ \beta_n \end{bmatrix} \tag{51}$$

From which we can find

$$\beta_n = a_q(n+1) + \sum_{k=1}^{n} v_n(k) a_q(n+1-k) \tag{52}$$

Equation (52) can be rewritten as Equation (54), as shown at the top of the next page, in which the direction of an array is flipped from up to down using the flipped method.

$$a_{n+2}*flipud(v_{n+1}) = \begin{bmatrix} \beta_n & 0 & 0 & \cdots & 0 & \varepsilon_n \end{bmatrix} \tag{53}$$

Combining Equation (51) and Equation (53) yields Equation (54). where we define

$$R_{n+1} = -\frac{\beta_n}{\varepsilon_n{}^*} \tag{55}$$

$v_{n+1}$ is determined as

$$v_{n+1}(k) = v_n(k) + R_{n+1} v_n{}^*(n-k+1); k = 0, 1, \ldots n+1 \tag{56}$$

we can have

$$SNR = \frac{r^{ACLDR}(0, y) - \mu^2}{r(0, y) - r^{ACLDR}(0, y)} \tag{57}$$

With error equation defined in Equation (59)

$$\varepsilon_{n+1} = \varepsilon_n \left[ 1 - |R_{n+1}|^2 \right] \tag{58}$$

$$a_{n+1} \left\{ \begin{bmatrix} 1 \\ v_n(1) \\ v_n(2) \\ \vdots \\ v_n(n) \\ 0 \end{bmatrix} + R_{n+1} \begin{bmatrix} 0 \\ v_n^*(n) \\ v_n^*(n-1) \\ \vdots \\ v_n^*(1) \\ 1 \end{bmatrix} \right\} = \begin{bmatrix} \varepsilon \\ 0 \\ 0 \\ \vdots \\ 0 \\ \beta_n \end{bmatrix} + R_{n+1} \begin{bmatrix} \beta^*n \\ 0 \\ 0 \\ \vdots \\ 0 \\ \varepsilon_n^* \end{bmatrix} a_{n+2} * flipud\,(v_{n+1}) = \begin{bmatrix} \beta_n & 0 & 0 & \ldots & 0 & \varepsilon_n \end{bmatrix} \quad (54)$$

### G. CUBIC HERMITE INTERPOLATION WITH LINEAR LEAST SQUARE REGRESSION SIGNAL-TO-NOISE RATIO (CHILLSRSNR) METHOD

Yeap et al. proposed the CHILLSRSNR technique in 2018 [58]. Equation (59) illustrates how the equation was created utilizing the concepts of Cubic Hermite Interpolation and Linear Least Square Regression [50].

$$S_i(x) = b_i(x - x_i) + c_i(x - x_i)^2 + d_i(x - x_i)^3 \quad (59)$$

Every interval of data points would have its own unique cubic function. Thus, the spline $S(x)$ is the function at data points. Four coefficients must be obtained, according to Equation (59). Equations (60) and (61) demonstrate how the spline ensures the exact occurrence of the data points.

$$S_i(x_i) = a_i \quad (60)$$

$$S_i(x_{i+1}) = S_{i+1}(x_{i+1}) = a_i \quad (61)$$

By using differentiation, Equations (62) and (63) are used to get equations (64) and (65), respectively, guaranteeing the smoothness of the $S(x)$. Equations (64), (66), and (67) must be solved in order to determine the coefficients (b, c, and d).

$$\frac{dS_i}{dx_{i+1}} = \frac{dS_{i+1}}{dx_{i+1}} \quad (62)$$

$$\frac{d^2S_i}{dx_{i+1}^2} = \frac{d^2S_{i+1}}{dx_{i+1}^2} \quad (63)$$

$$b_i + 2c_i x_{i+1} - 2c_i x_i x_{i+1} + 3d_i x_{i+1}^2 - 6d_i x_{i+1} x_i + 3d_i x_i^2 = 0 \quad (64)$$

$$2c_i - 2c_i x_i + 6d_i x_{i+1} - 6d_i x_i = 0 \quad (65)$$

$$S_i(x_{i+1}) = a_i + b_i(x_{i+1} - x_i) + c_i(x_{i+1} - x_i)^2 + d_i(x_{i+1} - x_i)^3 \quad (66)$$

The SNR is estimated using Equation (67). In this case, $S_i(0, y)$ represents the predicted noise-free peak at $x = (M + 1)/2$. In Equation (68), the noise peak is represented by h(0, y), and the mean is denoted by $\mu$.

$$SNR = \frac{S_i(0, y) - \mu^2}{h(0, y) - S_i(0, y)} \quad (67)$$

The next step is to formulate linear least square regression using Equation (69).

$$\frac{S_i(0, y) - \mu^2}{h(0, y) - S_i(0, y)} = R_i = \alpha r_i^2 + \beta r_i + \gamma_i + \varepsilon_i \quad (68)$$

In Matrix-vector form, Equation (69) can be rewritten as

$$R = XB + \varepsilon \quad (69)$$

where

$$R = \begin{bmatrix} R_1 \\ R_2 \\ . \\ R_N \end{bmatrix}, X = \begin{bmatrix} r_1^2 & r_1 & 1 \\ r_2^2 & r_2 & 1 \\ \vdots & \vdots & \vdots \\ r_N^2 & r_N & 1 \end{bmatrix}, B = \begin{bmatrix} \alpha \\ \beta \\ \gamma \end{bmatrix}, \text{ and } \varepsilon = \begin{bmatrix} \varepsilon_1 \\ \varepsilon_2 \\ . \\ \varepsilon_N \end{bmatrix}$$

The aim is to minimize the error term, thus we have

$$\left[ \min_\beta \sum_{k=1}^N \varepsilon_k^2 = \min_\beta \varepsilon^T \varepsilon \right] = \left[ \frac{d}{d\beta} \sum_{k=1}^N \varepsilon_k^2 = \frac{d}{d\beta} \varepsilon^T \varepsilon \right] = 0 \quad (70)$$

The minimization process is shown through the following steps. Let $\varepsilon^T \varepsilon = R - XB$. We have

$$\frac{d}{d\beta}\left(\varepsilon^T \varepsilon\right) = \frac{d}{d\beta}(R - XB)^T(R - XB) = 0$$

$$\frac{d}{d\beta}(R^T R - R^T XB + \left(XB\right)^T(XB) - (XB)^T R) = 0$$

$$\frac{d}{d\beta}\left(R^T R - R^T XB + B^T X^T XB - R^T XB\right) = 0$$

$$\frac{d}{d\beta}\left(R^T R - 2R^T XB + B^T X^T XB\right) = 0$$

$$2X^T XB - 2X^T R = 0$$

$$X^T XB = X^T R$$

And finally,

$$B = \left(X^T X\right)^{-1} X^T R = \begin{bmatrix} \alpha \\ \beta \\ \gamma \end{bmatrix}$$

Which then leads to

$$R_i = \alpha r_i^2 + \beta r_i + \gamma_i \quad (71)$$

where the estimated SNR is denoted by $R_i$, while the actual SNR is denoted by $r_i$.

### H. SUMMARY OF LIMITATIONS AND CONSTRAINTS

Traditionally, methods would require 2 images of the same sample area, and such practical constraints would include alignment problems as well as contamination risks, as well as the inability to support real-time applications. On the contrary, the existing SNR estimation techniques need only one image and are able to provide faster results. Techniques such as Nearest Neighbour (NN), First-Order Linear (FOL), their combination (NN+FOL), and advanced approaches, including LSR, NLLSR, ASNN, ACLDR and CHILLSRSNR provides different tunings in terms of speed, accuracy and stability compared to the traditional approaches such as Frank and Al-Ali method and SMART.

Table 2 summarizes the strengths and weaknesses of single image SNR estimation methods.

**TABLE 2.** Strengths and weaknesses of each single image SNR estimation methods.

| SNR estimation techniques | Strength | Weakness |
|---|---|---|
| NN Method | Fast and simple; good for initial SNR estimations. | One can underestimate SNR based on nearest-point approximation. |
| FOL Method | Offers improved estimation via interpolation. | It is commonplace for May to overestimate SNR due to a lack of precision in interpolation. |
| NN+FOL Method | Uses the strength of NN and FOL for an equal mode of output. | Shares limitations of both NN and FOL approaches. |
| LSR Method | Provides stable and accurate estimations. | Not effective with very non-linear SEM image data. |
| NLLSR Method | High accuracy in complex images by means of non-linear regression. | More computationally expensive than linear based methods. |
| ASNN Method | Consistent output with minimal fluctuations. | Poorer accuracy as a result of overly simplified modelling. |
| ACLDR Method | Good accuracy and stability and best for accurate analysis. | Computationally complex; less efficient for large datasets. |
| CHILLSRSNR Method | Balanced performance in executing cubic interpolation and linear regression. | Implicit, but may inherit limitations from component techniques. |

Besides the software-based SNR measurement, we can also apply SE yield to do the SNR measurement. In later section, we will discuss about the SNR measurement from hardware standpoint.

## VI. SNR MEASUREMENT USING HARDWARE

### A. SECONDARY ELECTRON YIELD SNR MEASUREMENT

In SEM analysis, standard images are generated by SEs, which result from inelastic interactions between the electron beam and conduction band electrons in the sample. The electron beam causes weakly bound conduction electrons to emit SEs, which typically have energies below 50 eV. SEs are excited by PEs and BSEs as they travel through the surface exit depth. The SE is excited by the PE and BSE trajectory through the surface exit depth. Thus, the SNR of SE [3], $SNR_{SE}$, is defined as:

$$SNR_{SE} = \bar{N}_{PE}\delta/[var(\bar{N}_{PE}\delta)]^{1/2} = [\bar{N}_{PE}/(1+b)]^{1/2}$$
$$= [I_{PE}/2e\Delta f(1+b)]^{1/2} \tag{72}$$

Fig. 13 shows a schematic setup for measuring SE and BSE yield, featuring an Everhart-Thornley (ET) detector [68]. The detector captures signals from BSEs, SE1 (SEs produced by PEs), SE2 (SEs generated by BSEs on the surface), and SE3 (SEs from BSEs interacting with the pole-pieces). High-energy electron paths are depicted as solid lines, while SE paths are shown as dotted lines. A manual switch allows easy reversal of the bias on the specimen holder and Faraday cup [33].

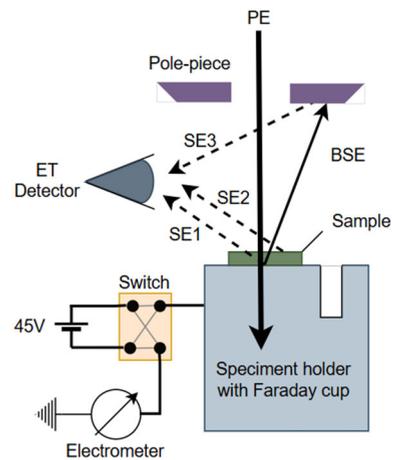

**FIGURE 13.** A schematic diagram which illustrates the setup for the SE and BSE yield measurement method as proposed in [29].

### B. THEORY AND MEASUREMENT TECHNIQUE

The experimental setup, including a combined specimen holder and Faraday cup, is shown in Fig. 13 and schematically in Fig. 14.

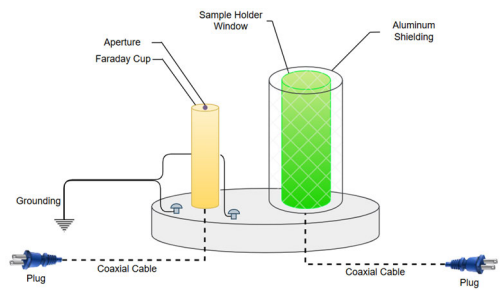

**FIGURE 14.** Imaging head for secondary-emission detection; the power supply and electrometer (not shown) are mounted externally to the SEM [33].

Unlike earlier techniques with exposed areas, this setup minimizes the exposure to SE3 [3]. Teflon insulation isolates the specimen stub from the motorized SEM stage. Fig. 14 depicts a simple manual switch that biases the holder at ±45V using five 9-V batteries (or alternatively, a ±50V DC power supply). The specimen holder has a diameter of 2.5 mm, and coaxial cables minimize electrical interference. A Keithley electrometer (Model 6512) measures the sample current ($I_{sc}$), primary electron current ($I_{PE}$), and voltage.

To measure $I_{PE}$, the Faraday cup is first positioned in the electron beam, and after removal, the sample replaces it. Measuring the specimen current ($I_{sc}$) with the holder biased at positive and negative voltages allows for the calculation of SE and BSE yields using the following relations as Equation (73).

$$I_{PE} = I_{SC} + I_{BSE} + I_{SE} \tag{73}$$

where, ISE represents the secondary electron (SE) current from all sources, and IBSE denotes the BSE current [4]. At zero bias, both SEs and BSEs are emitted from the specimen's surface. As shown in Fig. 15a, applying a negative bias ($-45V$) to the holder to ensure that all SEs and BSEs are repelled from the specimen. The measured specimen current ($I_{SC, -V}$) is then expressed by Equation (74)

$$I_{SC,-V} = I_{PE} - I_{BSE} - I_{(SE1+SE2)} \tag{74}$$

where $I_{SE1+SE2}$ is the current that produced from SE1 and SE2 (See Fig. 13). When the bias is reversed to +45V, as shown in Fig. 15b, low-energy SEs are attracted back to the specimen surface, while BSEs continue to reach the Everhart-Thornley (ET) detector.

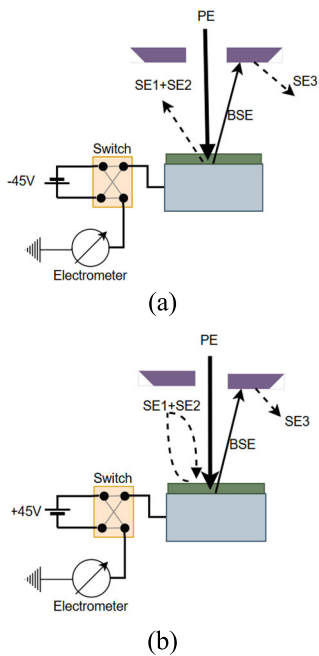

**(a)**

**(b)**

**FIGURE 15.** Electron trajectories under different DC bias voltages applied to the specimen holder under (a) negative bias, and (b) positive bias.

The resulting specimen current is expressed by Equation (75).

$$I_{SC,+V} = I_{PE} - I_{BSE} + I_{SE3} \tag{75}$$

where $I_{SE3}$ denotes the current that produced by SE3s from BSEs that hit the pole-piece. Subtracting Equation (74) from Equation (75) yields

$$I_{SE1+SE2} + I_{SE3} = I_{SC,+V} - I_{SC,-V} \tag{76}$$

Since the setup is prepared so that $I_{SE3}$ is less than $I_{SE1+SE2}$, the equation can be approximated as Equation

$$I_{SE} = I_{SE1+SE2} = I_{SC,+V} - I_{SC,-V} \tag{77}$$

Similarly, by adding Equation (74) and (75) and applying for the same approximation, we can have the relationship for backscattered electron current

$$I_{BSE} = I_{PE} - \frac{1}{2}(I_{SC,+V} + II_{SC,-V} + I_{SE1+SE2})$$
$$= I_{PE} - I_{SC,+V} \tag{78}$$

So, the secondary electron yield and back scattered electron yield are in terms of the measurables as

$$\delta = \frac{I_{SE}}{I_{PE}} = \frac{I_{SC,+V} - I_{SC,-V}}{I_{PE}} \tag{79}$$

$$\eta = \frac{I_{BSE}}{I_{PE}} = \frac{I_{PE} - I_{SC,+V}}{I_{PE}} \tag{80}$$

To validate this measurement technique, Sim and White in [33] measured the secondary electron yield ($\delta$) and backscattered electron yield ($\eta$) for several common materials, with the results compared to values obtained using more sophisticated methods in other studies. In each case, the sample surface was homogeneous. For instance, in the case of gold, a homogeneous surface was prepared by evaporating 99.9% pure gold onto a silicon wafer, resulting in a final thickness of approximately $\sim 4$ $\mu$m. The sample size was approximately $\sim 4$mm$^2$.

Fig. 16 and Fig. 17 show the measured values of backscattered electron and secondary electron yields, respectively, for Au, Si, Al, Cu, K, and In at 10 keV, 20 keV, and 30 keV by Sim in [33]. These results are compared with published databases, with the final column showing the differences between the values obtained in this study and those reported by Reimer [59], Bishop [60], Heinrich [61], Kanter [62], and Moncrieff [63]. For majority of measurements, the backscattered electron yield values vary by less than 10% from those published by Reimer and fall within the range of previously reported values. In contrast, the secondary electron yield ($\delta$) exhibits significant variation ($\sim 30\%$) between our data and those published by Reimer as well as among other published datasets. This discrepancy is attributed to the high sensitivity of $\delta$ to surface conditions, underscoring the importance of in-situ measurements rather than relying solely on published values.

Although electrons released by PE exhibit a range of energies, the majority have either very low or very high energies, resulting in a double-peaked distribution. Electrons with energies E<50eV are arbitrarily classified as secondary electrons, while those with E>50eV, elastically scattered or Auger electrons) are categorized as backscattered electrons [64].

In the measurements presented here, the voltage was alternated between ±45V to simplify the setup. To assess the potential error introduced by using ±45V instead of ±50V, measurements on gold and silver were repeated with a switchable power supply. The difference in calculated SE and BSE

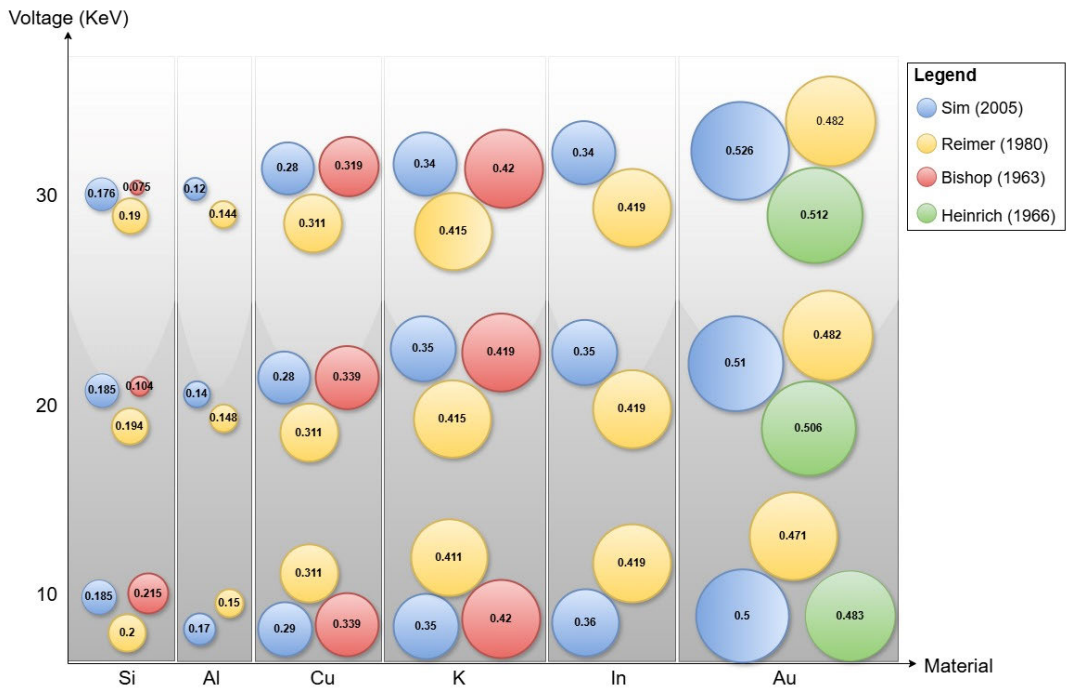

**FIGURE 16.** In-situ measurement of backscattered electron yield ($\eta$) for various elements as a function of energy as conducted and presented in [33] and compared with published data. The sample holder was biased at $\pm 45V$ for these measurements, while the E-T detector cage was biased at $-150V$ to ensure that the E-T detector would not compete for SEs with the voltage biasing at the specimen holder.

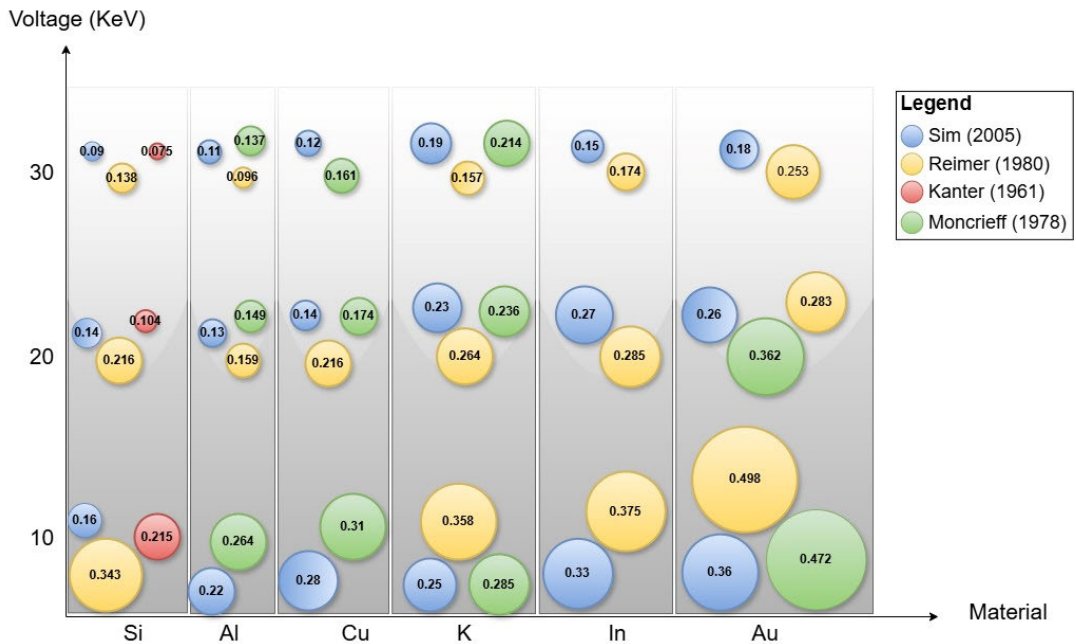

**FIGURE 17.** In-situ measurement of secondary electron yield ($\delta$) for various elements as a function of energy as conducted and presented in [33] and compared with published data. The sample holder was biased at $\pm 45V$ for these measurements, while the E-T detector cage was biased at $-150V$ to ensure that the E-T detector would not compete for SEs with the voltage biasing at the specimen holder.

yields when using $\pm 45V$ rather than $\pm 50V$ was found to be less than 2.5% in all cases.

In the previous section, the approximation $I_{SE3} \ll I_{SE1+SE2}$ was used. To verify the validity of this assumption,

a simple experiment was conducted. To block SE3, a 2 $\times$ 2-inch Aluminum (Al) plate was coated with approximately 15 $\mu$m of carbon paint. Given that the Kanaya-Okayama range for carbon is 5.3 $\mu$m at 20 keV and 10.4 $\mu$m at 30 keV (as noted in Table 3.2 of [4]), this coating thickness is sufficient to absorb $I_{SE3}$. Additionally, the Bethe range is 7.5 $\mu m$ and 13 $\mu m$ for 20 keV and 30 keV, respectively, according to Goldstein [4].

The specimen current was obtained both with and without the Aluminum (Al) plate covering the pole-piece of the SEM. The summary for the experiments using gold and silicon samples are presented in Table 3. For the bright gold sample, SE3 contributes less than 3% to the total secondary electron (SE) current, while for the darker silicon sample, the contribution is approximately 1%. These results suggest that SE3 is negligible across all sample types, validating the approximations used to derive the secondary electron and backscattered electron yields within the experimental setup, which minimizes exposure to SE3.

Another potential source of measurement error is current leakage during the determination of $\delta$ (and $\eta$). However, this error should be minimal because:

1. Proper cable shielding has been implemented to reduce current leakage.

2. The form of the equation used to determine $\delta$ ensures that any consistent underestimation of specimen current will, to a first-order approximation, cancel out.

### C. PRIMARY ELECTRON YIELD SNR MEASUREMENT

The SNR in a SEM is determined by the combined contributions of PE, SE, and BSE. In SEMs equipped with thermionic electron guns, shot noise in the PE beam is the predominant noise source [28] and it adheres to Poisson statistics [3]. For BSEs, although the conversion from PEs to BSEs follows binomial distribution, the integration of the Poisson statistics of PEs with binomial conversion results in BSE emissions also exhibiting a Poisson distribution.

In the subsequent discussion, Reimer's derivation [3] is used to calculate the signal-to-noise ratio (SNR) by using the parameters $I_{PE}$, $\delta$, $\eta$, and the acquisition time per pixel of the digital image ($\tau$). The mean number of primary electrons per pixel is expressed as: $\bar{N}_{PE} = I_{PE}t/e$, where $e$ is the charge of a single electron. The SNR of primary electrons is then expressed as

$$SNR_{PE} = \bar{N}_{PE}/[var(N_{PE})]^{1/2} = \bar{N}_{PE}^{1/2} \quad (81)$$

For backscattered electrons, the cascade of the Poisson distribution of PE and the binomial distribution of the conversion factor $\eta$ yields

$$SNR_{BSE} = \bar{N}_{PE}\eta/[var(\bar{N}_{PE}\eta)]^{1/2} = (\bar{N}_{PE}\eta)^{1/2} \quad (82)$$

However, the noise contributed by secondary electrons cannot be as simply modelled. The SE distribution is neither Poisson nor binomial since a single primary electron can release zero, one, or multiple secondary electrons with decreasing probability [4], [65]. The SNR for secondary electrons is given by

$$SNR_{SE} = \bar{N}_{PE}\delta/[var(\bar{N}_{PE}\delta)]^{1/2} = [\bar{N}_{PE}/(1+b)]^{1/2} \quad (83)$$

where $b = var(\delta)/\delta^2$. In the limiting case of Poisson statistics, $var(\delta) = \delta$ and consequently, $b = 1/\delta$. Deviations from Poisson statistics increased by a factor of 1.2 (Al) to 1.5 (Au), depending on the material and electron energy.

Using the known values of $I_{PE}$ and $\tau$, and the measured values of $\eta$ or $\delta$, the SNR for images based on backscattered or secondary electrons can be determined. For secondary electrons, material-dependent noise enhancement must also be included.

Finally, the SNR measured by the Everhart-Thornley (E-T) detector is lower than that derived from the incident electron dose and electron yield factor due to the finite detector quantum efficiency (DQE). The relationship is given by [66]

$$SNR_{ET} = \sqrt{DQE}SNR_{yield} \quad (84)$$

The DQE or collection efficiency only needs to be measured once for a given microscope/detector. While the methodology is adequately described in the literature (Joy et al., [43]), it is beyond the scope of this paper. We estimate the DQE of our instrument to be 0.23, which is within the published range of 0.15 to 0.25 ([43], [66]).

The calculated SNR for images generated by secondary electron emission for silicon and gold specimens is summarized in Tables 4 and 5, respectively, as a function of electron energy. The SNR values remain remarkably constant over the voltage range studied, despite a 50% reduction in secondary electron yield at higher incident energies.

In experimental work, the SNR of an image is often estimated directly from the image using the relationship

$$SNR = (I_{mean} - I_{DC})/\sigma \quad (85)$$

where $I_{mean}$ represents the mean intensity of the image averaged over all pixels, $\sigma$ is the standard deviation of the intensity recorded at each pixel, and $I_{DC}$ is the mean intensity of the image averaged over all pixels at zero beam current.

Experimentally, $I_{DC}$ is obtained during system calibration prior to performing digital imaging. This calibration is essential to ensure that measurement results do not saturate either the upper or lower limits of the video dynamic range. A convenient procedure to achieve this calibration and simultaneously obtain $I_{DC}$ involves controlling the incident beam current using the SEM aperture.

The calibration process begins by adjusting the SEM aperture to minimize the signal as closely as possible to zero, followed by capturing an image. The aperture is then sequentially switched from the largest to the smallest available diameter over a few seconds during the scan, with the incident beam current recorded at each setting. From the resulting image, the mean intensity corresponding to each beam current can be calculated.

Fig. 18 shows plots of $I_{mean}$ as a function of beam current for both Gold (Au) and Silicon (Si) samples. The slope of these plots varies between the samples, with the mean intensity for Au showing a greater dependence on beam current. Both plots exhibit linear behavior, with no evidence of saturation at the given primary electron current. The $I_{DC}$ values, obtained from the y-axis intercepts, are 12.3 and 12.1 for Au and Si, respectively.

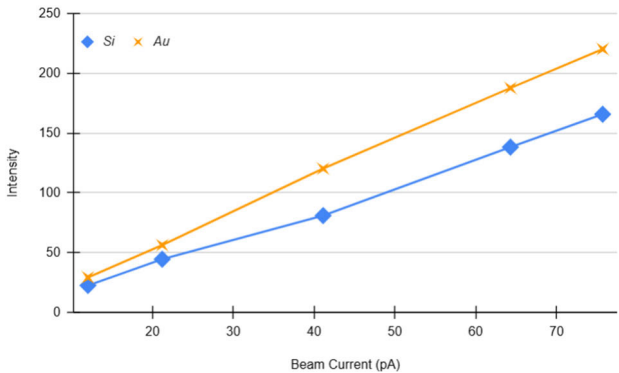

**FIGURE 18.** Mean image intensity as a function of beam current for gold (Au) and silicon (Si) samples at a primary electron current IPE of 379 pA and an accelerating voltage of 20 keV. Experimental data points are represented by stars for Au and diamonds for Si. The y-axis intercepts, indicating the mean intensity at zero beam current, are 12.3 for Au and 12.1 for Si, respectively [33].

Table 3 and Table 4 compare the SNR of images based on secondary electrons, calculated using the secondary electron yield, with that estimated from the image itself for silicon and gold at three different voltages each. A comparison of the calculated and estimated SNR reveals that there is significantly greater variation with electron energy in the SNR estimated from the images than in the SNR calculated based on electron yield.

Furthermore, the difference between the averaged SNR values (across electron energy) is approximately 10% for the silicon specimen and around 20% for the gold specimen. Several factors may contribute to these deviations. First, it is possible that the enhancement of b due to non-Poisson statistics has been overestimated. If Poisson statistics are assumed, then the two calculation methods agree within 5%.

Second, while SE3s do not affect the SE yield measurement, they may influence the image SNR. Third, for an ideal SE detector, 100% of the emitted secondary electrons would reach the detector, while no BSEs would be detected. In practice, however, a small number of BSEs is detected, which may either positively or negatively affect the SNR.

**TABLE 3.** SNR derived from secondary electron yield vs estimation from image of homogeneous silicon specimen.

| Voltage [keV] | SNR based on electron yield | | | | SNR from image | | | |
|---|---|---|---|---|---|---|---|---|
| | $\delta$ | $I_{PE}$ [pA] | $SNR_{SE}$ | $SNR_{ET}$ | $I_{DC}$ | $I_{mean}$ | $\sigma$ | $SNR_{ET}$ |
| 10 | 0.16 | 291 | 40 | 19 | 12.1 | 68 | 2.35 | 24 |
| 20 | 0.14 | 371 | 42 | 20 | 12.3 | 45 | 1.72 | 19 |
| 30 | 0.09 | 495 | 40 | 19 | 11.8 | 41 | 1.35 | 22 |

The results as presented in Table 2 and Table 3 may lead to the conclusion that the SNR estimated from the image is always equivalent to that calculated using electron yields.

**TABLE 4.** SNR derived from secondary electron yield vs estimation from image of homogeneous gold specimen.

| Voltage [keV] | SNR based on electron yield | | | | SNR from image | | | |
|---|---|---|---|---|---|---|---|---|
| | $\delta$ | $I_{PE}$ [pA] | $SNR_{SE}$ | $SNR_{ET}$ | $I_{DC}$ | $I_{mean}$ | $\sigma$ | $SNR_{ET}$ |
| 10 | 0.16 | 291 | 40 | 19 | 12.1 | 68 | 2.35 | 24 |
| 20 | 0.14 | 371 | 42 | 20 | 12.3 | 45 | 1.72 | 19 |
| 30 | 0.09 | 495 | 40 | 19 | 11.8 | 41 | 1.35 | 22 |

While this is generally true for a perfectly homogeneous specimen surface, any real contrast in the image will lower the apparent SNR [67]. For example, in a high-contrast image, $\sigma$ will be large, leading to an artificially low SNR value [24]. Calculating SNR from electron yield eliminates this convolution.

This section has demonstrated that knowledge of the primary electron current and specimen current, with the sample holder biased at $\pm 45$V, is sufficient to determine both the secondary electron yield ($\delta$) and backscattered electron yield ($\eta$).

### D. FACTORS WHICH AFFECT SNR MEASUREMENT
In the following section, we would like to discuss about the various aspects that can affect the SNR measurement for the SEM.

#### 1) ACCELERATING VOLTAGE
In this section, sample images of the mold compound in a power IC package are presented. Three images, taken at accelerating voltages ranging from 10 keV to 30 keV, are presented in Fig. 19. The image contrast can be enhanced by adjusting the accelerating voltage, as noted by [8], [68] and [73]. Lowering the accelerating voltage increases the beam diameter, which in turn impacts the image contrast.

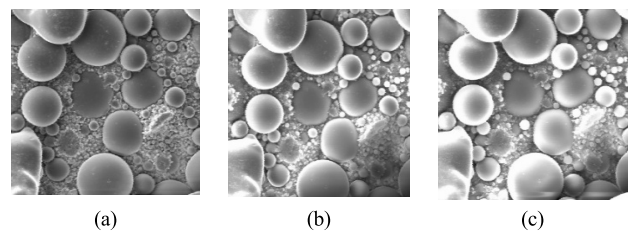

**FIGURE 19.** Sample images of mold compound of power IC package captured at (a) 10 keV, (b) 20 keV and (c) 30 keV. Horizontal field-width = 50 $\mu$m.

The ACF curves shown in Fig. 20 show how the contrast of the image is affected by the accelerating voltage.

#### 2) BEAM DIAMETERS
In this section, the influence of the beam diameter in SEM imaging is examined. which directly affects the shape of the ACF curve. The probe size of the electron beam plays a crucial role in determining resolution. Generally, reducing the probe size enhances the spatial resolution of the image. However, the probe size is influenced by several factors:

a) **Accelerating Voltage**: Lowering the accelerating voltage increases the probe size; while raising it reduces the wavelength of the electron beam, resulting in a smaller probe size.

b) **Condenser Lens Current**: Increasing the current in the condenser lens reduces the focal length of the crossover produced between the condenser lens and the objective lens, leading to a smaller probe size.

c) **Working Distance**: Reducing the working distance decreases spherical aberration, thereby producing a probe size with a smaller diameter.

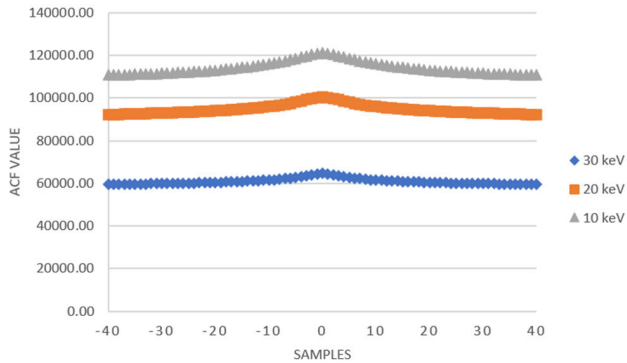

**FIGURE 20.** The ACF curves of sample images of mould compound of power IC package captured at various accelerating voltage from 10 keV to 30 keV.

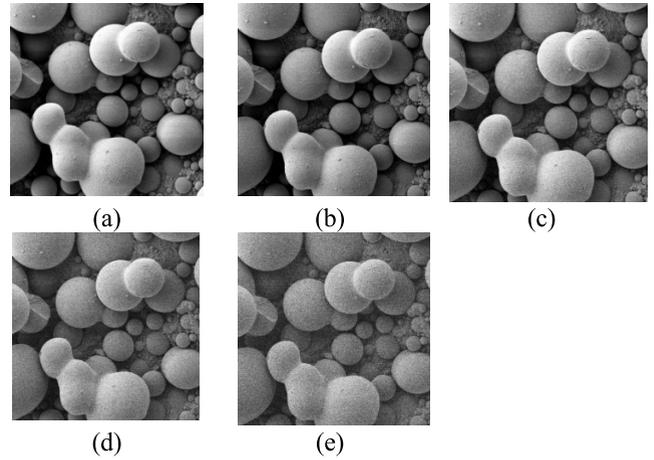

**FIGURE 21.** Sample images of power IC captured at (a) beam diameter = 151 nm, (b) beam diameter = 89 nm, (c) beam diameter = 60 nm, (d) beam diameter = 25 nm and (e) beam diameter at 18 nm. Image size = 256 × 256 pixels and beam energy = 5 keV.

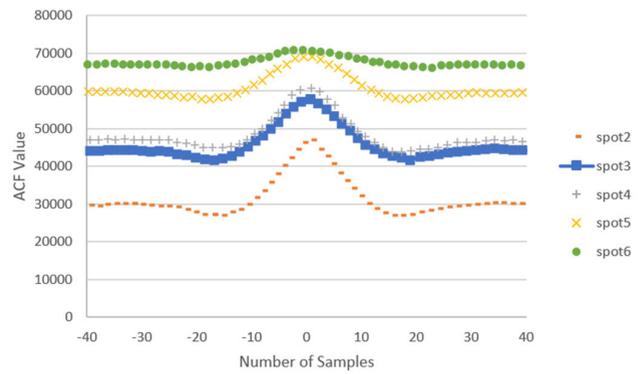

**FIGURE 22.** The ACF curves of sample images of mold compound of power IC package captured at various beam diameters from 151 nm to 25 nm.

For larger beam diameters, improved image contrast is typically achieved. Conversely, smaller beam diameters result in images with finer detail but may introduce more noise.

In this experiment, the cell of a power transistor package was used as the sample. Images were captured at beam diameters ranging from 151 nm to 25 nm, as detailed in Fig. 21. The corresponding ACF curves for various beam diameters are shown in Fig. 22. These results indicate that smaller beam diameters result in poorer SNR, whereas larger beam diameters yield better SNR performance, as summarized in Table 5.

### 3) SAMPLES WITH AND WITHOUT HEAVY METAL COATING
The thickness of a metal coating can influence the shape of the ACF curve. In this experiment, we used a sample image of a power transistor package. Initially, the sample was coated with gold to a thickness of 91 Å. After the first round of sample viewing, the same sample was re-sputtered to a coating thickness of 1091 Å.

After performing the ACF analysis, the ACF curves for various coating thicknesses were plotted and are shown in Fig. 23. The results indicate that a thicker coating leads to a better SNR. The changes in the ACF curve from 91 Å to 1091 Å are depicted in Fig. 23.

Please note that the data considered a smooth, homogeneous surface. If the sample surface is rough, the results would differ. The ideal coating thickness depends on the roughness of the sample. Once continuous conductive coverage is achieved, increasing the coating thickness doesn't improve the results further. For a smooth sample, a thin coating may be sufficient, while for a rough sample, a thicker coating is needed to ensure proper conductivity in recessed areas. However, in such cases, the coating may vary in thickness, leading to potential artifacts [15]. A 1 − μm gold coating, for example, would completely obscure the surface details.

**TABLE 5.** Experimental results for power transistor package sample image captured at beam diameter.

| Beam Diameter (nm) | Noisy ACF | Noise free | Mean | Signal | Noise | SNR | SNR (dB) |
|---|---|---|---|---|---|---|---|
| 151 | 77279.9 | 77251 | 75289.8 | 1961.2 | 28.9 | 67.86 | 18.32 |
| 89 | 77177.3 | 77149.6 | 75331.4 | 1818.2 | 27.7 | 65.64 | 18.17 |
| 60 | 77114.6 | 77070.3 | 75323 | 1747.3 | 44.3 | 39.44 | 15.96 |
| 38 | 77050.6 | 76991 | 75227.2 | 1763.8 | 59.6 | 29.59 | 14.71 |
| 25 | 76591 | 76538.5 | 75990.3 | 548.2 | 52.5 | 10.44 | 10.18 |

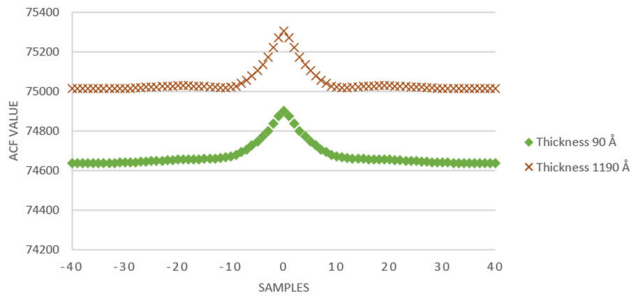

#### 4) SURFACE TILT

To investigate surface tilt, we used the sample image of the power IC package cell. Images were captured at tilt angles of 0°, 10°, and 20° (see Fig. 24). After performing the ACF analysis, the resulting curves for each tilt angle are shown in Fig. 25. As the tilt angle increases toward the SEM detector, the signal for the SNR also increases, leading to an enhancement in the ACF from 0° upwards.

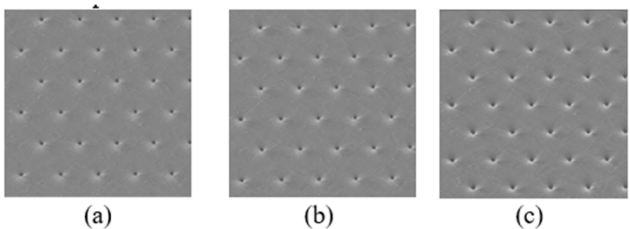

(a)          (b)          (c)

**FIGURE 24.** The ACF curves of sample images of the cell of power IC package captured at various angle of tilt of (a) 0°, (b) 10°, and (c) 20°. Horizontal field-width = 20 $\mu$m and beam energy = 20 keV.

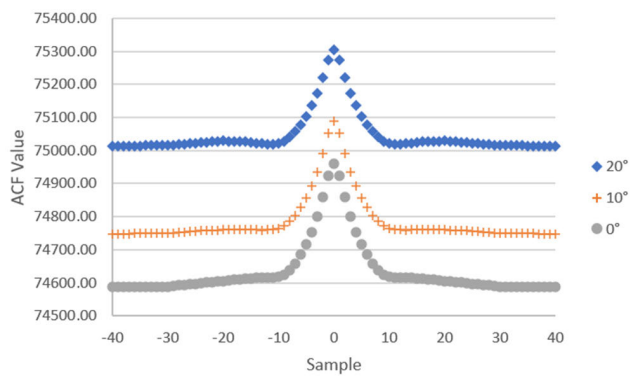

**FIGURE 25.** The ACF curves of sample images of the cell of power IC package captured at various angles of tilt from 0° to 20°.

#### 5) UNDER-SAMPLING

Under sampling is the rate of scan fail to capture and record higher frequency components of the specimen. The Nyquist rate system is commonly used in signal digitization [69], where the sampling frequency is at least twice the maximum signal frequency of interest ($f_{max}$) In over-sampling, the sampling frequency exceeds the basic Nyquist rate of $2f_{max}$.

When images of a specimen are captured at an under-sampling rate, the specimen's spatial components at frequencies higher than the Nyquist frequency are not recorded. Without prior knowledge of the specimen, it is challenging to determine whether the image has been captured in under-sampling conditions. The ACF curve is derived from the statistical information of the targeted area. However, restoring the unrecorded information is difficult without sufficient statistical data.

#### 6) CONTRAST IN SEM IMAGES

One way to modify the contrast of an image is by adjusting the accelerating voltage. Reducing the accelerating voltage improves image contrast, as it increases the beam diameter, which in turn enhances the contrast. Fig. 26 illustrates the impact of varying the accelerating voltage on image contrast.

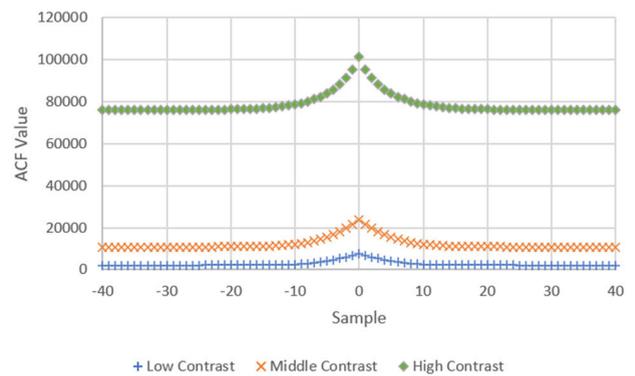

**FIGURE 26.** The ACF curves of sample images of mould compound of power IC package captured at various contrasts.

#### 7) SCANNING RATE OF SEM

SEMs are often operated at the television (TV) rate doing instrument adjustment and study of dynamics [40]. However, the TV rate electron microscope images invariably have poor SNR owing to the small number of electrons making up each frame; it is desirable to have some means of processing the video signals to reduce the noise. In SEM, it provides several scanning rates and varies from the TV-rate display to slow scan rate. In this section, we verify that the slower the scan rate, the better the quality in term of SNR value of the SEM image. Images in Fig. 27 are those captured from TV scan rate to various slow-scan rates. The number of scanning rate, is gradually increased, as shown as in Fig. 27. From Fig. 27h, SEM has the scanning rate of 160 seconds per frame. The details of SNR and scan rate can be referred to Table 6.

### VII. MODERN APPROACHES FOR SNR ESTIMATION
#### A. MACHINE LEARNING-BASED ESTIMATION

Recent advancements in machine learning (ML) have led to the development of data-driven methods for SNR estimation, leveraging deep learning architectures like Convolutional Neural Networks (CNNs) to analyze large datasets of SEM images. These models can learn the noise characteristics from the data and estimate SNR in real-time with high accuracy, even in complex, non-stationary noise environments.

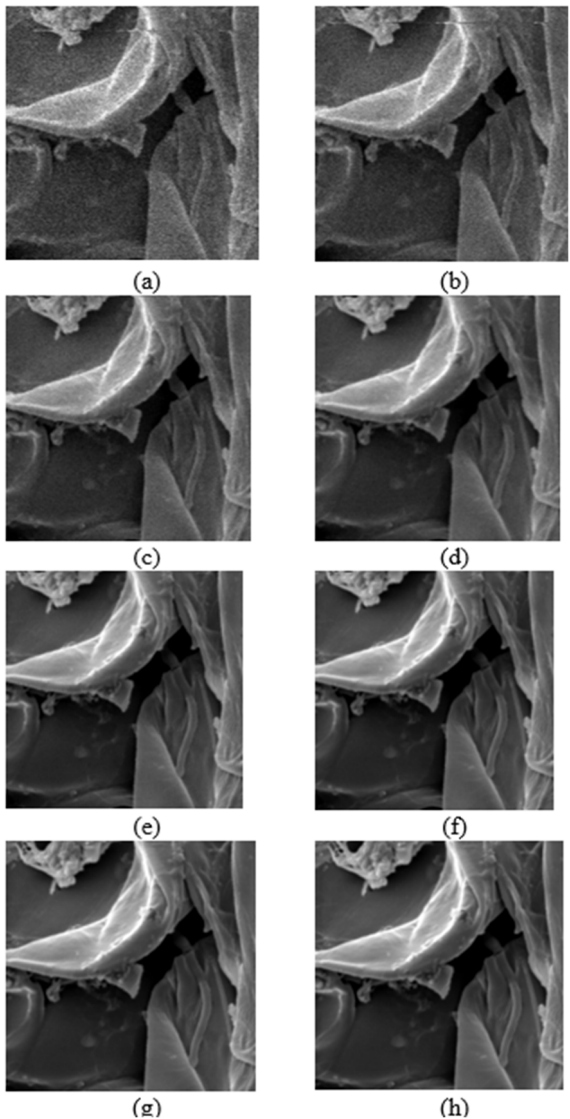

(a)   (b)

(c)   (d)

(e)   (f)

(g)   (h)

**FIGURE 27.** Scan rate of Composite nano material at various speed (a) TV scan rate, (b) Slow scan 1, (c) Slow scan 2, (d) Slow scan 3, (e) Slow scan 5, (f) Slow scan 5, (g) Slow scan 6 and (h) Slow scan 7. Image size = 256 × 256 pixels. Horizontal field width = 50 μm and beam energy = 16 keV.

**TABLE 6.** The SNR of bamboo images taken at various scan rate.

| Image | Scan Rate (Second) | SNR |
|---|---|---|
| TV scan rate | 0.033 | 11.74 |
| Slow scan 1 | 1 | 18.48 |
| Slow scan 2 | 2 | 41 |
| Slow scan 3 | 6.48 | 77 |
| Slow scan 4 | 23.8 | 75 |
| Slow scan 5 | 58.4 | 76 |
| Slow scan 6 | 81.6 | 74 |
| Slow scan 7 | 159.55 | 75.7 |

Machine learning, especially deep learning, has shown significant promise in automating and improving SNR estimation across various imaging modalities, including SEM. Deep learning models, particularly CNNs, have been widely used in tasks like noise reduction, image segmentation, and feature extraction. Early research highlighted the potential of CNNs in image recognition, which paved the way for their application in SNR estimation. Their hierarchical learning structure allows them to distinguish noise from signal effectively, especially in high-resolution images. Deep learning architectures, trained on large datasets of labelled images, have significantly improved image quality and enhanced signal detection by identifying inherent patterns and features in the data.

Several studies have explored the use of machine learning to enhance SEM image quality. For example, a deep learning model tailored for SEM images, incorporating domain-specific knowledge, has been developed to improve SNR estimation accuracy. This model combined CNNs and recurrent neural networks (RNNs) to capture both spatial and temporal correlations in the image data, showing notable improvements over traditional methods, particularly in high-noise environments. The integration of ML models into SEM systems has enabled the creation of automated workflows for image acquisition and analysis. Additionally, a Non-Data-Aided (NDA) SNR estimator, based on deep learning, has been proposed and shown to be effective for both baseband and intermediate-frequency signals when compared with traditional methods such as the M2M4 estimator.

Another promising deep learning approach is the Gaussian-Noise Convolutional Neural Network (GN-CNN), designed to classify noise variance in SEM images, which is crucial for enhancing image quality. The GN-CNN architecture includes layers such as an encoder layer with bi-directional LSTM, a convolutional layer with ResNet34, an attention layer, a decoder with LSTM cells, and a decision layer. This sophisticated design allows the model to classify SEM images into multiple noise variance categories with high accuracy (93.8%).

Moreover, deep learning models like Cycle-consistent Generative Adversarial Networks (CycleGAN) have been used to enhance SEM image quality, particularly for challenging samples with poor conductivity. CycleGAN, a form of unsupervised learning, improves image quality by assessing the Peak Signal-to-Noise Ratio (PSNR), which measures the restoration of image details. Higher PSNR values indicate better image restoration.

Despite these advancements, there are still challenges in the widespread adoption of machine learning for SNR estimation in SEM images. The accuracy and generalizability of ML models depend heavily on the quality and diversity of the training data. To ensure that models perform well across various samples and imaging conditions, comprehensive datasets and robust validation procedures are essential. Additionally, the interpretability of ML models is crucial for building user trust and enabling smooth integration into existing SEM workflows.

### B. BAYESIAN ESTIMATION

Bayesian approaches treat SNR estimation as an inference problem, where the goal is to estimate the signal and noise components based on prior knowledge and observed data. These methods are particularly effective when noise is non-Gaussian or when the signal is highly variable across the image.

## VIII. NOISE REDUCTION TECHNIQUES FOR IMPROVING SNR

Improving SNR in SEM images is a multi-faceted challenge that requires both hardware and software-based solutions [33]. Below, we discuss the key approaches to noise reduction in SEM.

### A. HARDWARE-BASED SOLUTIONS

Hardware solutions focus on reducing noise at the source or during signal acquisition.

#### 1) OPTIMIZING SEM PARAMETERS

One of the simplest methods for improving SNR is to adjust SEM operating parameters such as the electron beam current, accelerating voltage, and working distance. Higher beam currents result in more electrons reaching the detector, thereby increasing the signal and reducing the impact of shot noise. However, increasing the current may also damage sensitive samples or introduce other artifacts.

#### 2) ADVANCED DETECTOR TECHNOLOGY

The development of advanced detector systems, such as low-noise amplifiers and high-sensitivity detectors, has also contributed to SNR improvement. These technologies minimize the introduction of electronic noise during signal acquisition, leading to clearer images.

### B. SOFTWARE-BASED SOLUTIONS

Software solutions focus on reducing noise after the images are acquired.

#### 1) LINEAR FILTERING TECHNIQUES

Traditional linear filters, such as Gaussian filters and Wiener filters, have been widely used to reduce noise in SEM images. Gaussian filters smooth the image by averaging pixel values in a neighborhood, effectively reducing high-frequency noise but at the cost of blurring fine details. Wiener filters, on the other hand, are designed to minimize the mean square error between the filtered image and the original signal, making them more effective for images with varying noise levels.

#### 2) NON-LINEAR FILTERING TECHNIQUES

Non-linear filtering methods, such as median filters and bilateral filters, have been developed to address the limitations of linear filters. Median filters replace each pixel value with the median of its neighborhood, preserving edges while removing noise. Bilateral filters extend this concept by considering both spatial and intensity differences between pixels, allowing for effective noise reduction without significant loss of detail.

#### 3) WAVELET-BASED DENOISING

Wavelet transforms have gained popularity in SEM image processing due to their ability to decompose an image into multiple scales and orientations. This allows for selective noise removal in the wavelet domain. By thresholding the wavelet coefficients, noise can be effectively reduced while preserving important image features such as edges and textures.

#### 4) MACHINE LEARNING-BASED DENOISING

Machine learning has opened new avenues for SEM noise reduction. CNNs trained on large datasets can learn to differentiate between noise and signal, enabling real-time denoising with minimal human intervention. These methods have shown remarkable performance in preserving fine details while reducing noise, making them ideal for high-resolution SEM images.

## IX. NOISE REMOVAL TECHNIQUES

Noise in SEM images is a rather difficult issue to handle. The signal-to-noise ratio (SNR) of the images depends on the beam current, the materials present in the specimen, and the beam topography [3].

SEMs are often operated at television rate (TV) when adjusting for instrument. The TV rate SEM images have poor SNR due to fewer numbers of electrons making up each frame. Catto and Smith in [70] suggested using an image storage tube. However, this method cannot produce continuous noise reduced output. Many researchers have also developed digital techniques for SEM image analysis and processing ([2], [71], [72], [73]) to improve the quality of images. Boyes et al. [74] generated useful methods for the acquisition, analysis, processing, digital storage and correction of images.

In the late 1970's, a few techniques for noise removal in SEM ([75], [76], [77]) were introduced. Frank and Al-Ali in [39] proposed the idea of comparing variance and covariance in image. By using the concept of hysteresis smoothing, Duda and Hart [78] and Ehrich [79] proposed different idea of noise removal technique. However, due to the extremely severe processing artifacts, the hysteresis smoothing technique has not been introduced to the SEM field.

In 1982, Smith discussed about the limitation of two modes in the SEM imaging system [73]: the averaging mode and the recursive mode. In the same year, Eramus [40] recommended Kalman filtering by combining the efficiency of the averaging mode with some of the convenience of the recursive mode. The Kalman filtering is then able to reduce the noise in TV rate electron microscope images.

Oho et al. [80] developed a smoothing filter, the complex hysteresis smoothing (CHS) technique for noise removal of SEM without artifacts. In 2004, Oho et al. used a covariance and variance to choose the best focused image from a series of SEM images by changing the focus of objective lens under noisy SEM image condition.

Across various imaging domains, quantifying noise variance is vital for multiple reasons. To begin with, this metric serves as an indicator of image quality—elevated noise variance can obscure subtle signal information, making fine image details harder to discern. Moreover, having an accurate estimate of noise variance is essential when using algorithms that require it as an input parameter. Examples include Wiener filtering for image restoration [81], smoothing with Kalman filters [82], and optimal threshold selection [83] are among the applications.

Canny [84] estimated noise variance by isolating signal from noise within the outputs of his edge detector. He assumed that the lowest 80% of response amplitudes in the histogram represented noise and used this subset for his variance calculation. Voorhees and Poggio [85] approached the problem by modeling the distribution of image gradient magnitudes with a Rayleigh distribution fitted to the histogram. Meanwhile, Meer et al. [86] leveraged the orthogonality of their template-based edge masks to derive a noise variance estimate from image regions classified as uniform. In 1996, Venturini et al. [87] analyzed the digital image quality based on the Wiener technique. An adaptive Wiener filtering approach, based on image estimation in the wavelet domain, was proposed by Stephanakis et al. [88].The gradient-based estimation of the image was employed by minimizing an error function that depends on estimates of the image and the power of the noise in each wavelet sub-band. The power of the noise was then estimated from the variance of wavelet coefficients. Recently, Yamane et al. in [89] proposed an adaptive Wiener filter (AWF) based on the Gaussian mixture distribution model (GMM) as a realization of an optimum restoration filter.

Nevertheless, For the majority of the aforementioned techniques, the estimations' accuracy is unavailable. Serious constraints are introduced by their reliance on the noisy image's edge and/or uniform patches as discriminating features before estimate. Edge detection and surface fitting are no longer robust procedures at low SNR values (noise is comparable with picture fluctuation), and the regions differentiated as uniform or the edge detector magnitude histogram may not be trustworthy.

With the availability of high-performance Personal Computer (PC) and low running cost of PC ([80], [90], [91]) the possibility to obtain real-time signal-to-noise ratio to quantify SEM images can be achieved. However, there has not been any work done on the real-time noise removal technique in the SEM images although many noise removal techniques have been reported [57], [92], [93], [94], [95].

## A. WIENER FILTERING
A Wiener filter [96] is the mean square error (MSE)-optimal stationary linear filter for images degraded by additive noise and blurring. Calculation of the Wiener filter requires the assumption that the signal and noise are both second-order moments. For this description, only noise processes with zero mean will be considered without loss of generality.

The aim of the Wiener filtering is to estimate the original signal from a degraded version of the signal. The degraded image $w(n_1, n_2)$ is shown as Equation (86).

$$w(n_1, n_2) = f(n_1, n_2) + u(n_1, n_2) \quad (86)$$

where $f(n_1, n_2)$ is the noise free image and $u(n_1, n_2)$ is the noise.

Given the degraded image $w(n_1, n_2)$, a function $h(n_1, n_2)$ that can provide a good estimate of $f(n_1, n_2)$ is generated. This estimate is $z(n_1, n_2)$ and defined as in Equation (87).

$$z(n_1, n_2) = w(n_1, n_2) * h(n_1, n_2) \quad (87)$$

where (*) refers to convolution operation. The Wiener filter generates $h(n_1, n_2)$ that minimizes the mean square error VOLUME XX, 2017 7 (MSE) between the degraded image and noise-free image. The MSE is then defined as Equation (88)

$$E\left\{e^2(n_1, n_2)\right\} = E\left\{(f(n_1, n_2) - z(n_1, n_2))^2\right\} \quad (88)$$

where $error = e(n_1, n_2) = f(n_1, n_2) - z(n_1 - n_2)$ and $E$. denotes the expected value. The goal is to minimize the mean squared error between $z(n_1, n_2)$ and $f(n_1, n_2)$. Applying the principle of orthogonality, we have Equation (89).

$$E\left\{e(n_1, n_2)w^*(m_1, m_2)\right\} = 0, \ \forall(n_1, n_2), (m_1, m_2). \quad (89)$$

Thus,

$$\begin{aligned} E\left\{f(n_1, n_2)w^*(m_1, m_2)\right\} \\ = E\left\{(e(n_1, n_2) + z(n_1, n_2))w^*(m_1, m_2)\right\} \\ = E\{e(n_1, n_2)w^*(m_1, m_2) + z(n_1, n_2)w^*(m_1, m_2)\} \end{aligned} \quad (90)$$

Since $E\left\{e(n_1, n_2)w^*(m_1, m_2)\right\} = 0$ and $z(n_1, n_2) = w(n_1, n_2) * h(n_1, n_2)$, Equation (90) becomes

$$\begin{aligned} E\left\{f(n_1, n_2)w^*(m_1, m_2)\right\} \\ = E\left\{(w(n_1, n_2) * h(n_1, n_2))w^*(m_1, m_2)\right\} \\ = \sum_{i_1=-\infty}^{\infty}\sum_{i_2=-\infty}^{\infty} h(i_1, i_2) \\ E\{w(n_1 - i_1, n_2 - i_2)w^*(m_1, m_2)\} \end{aligned} \quad (91)$$

Equation (91) can be rewritten as

$$\begin{aligned} R_{fw}(n_1 - m_1, n_2 - m_2) \\ = \sum_{i_1=-\infty}^{\infty}\sum_{i_2=-\infty}^{\infty} h(i_1, i_2)R_w(n_1 - i_1 - m_1, n_2 - i_2 - m_2) \end{aligned} \quad (92)$$

where $R_{fw}$ refers to the cross correlation between the noise-free and noisy images and $R_w$ refers to the autocorrelation of the noisy images.

Thus, Equation (92) can be reformulated as

$$R_{fw}(n_1, n_2) = h(n_1, n_2) * R_w(n_1, n_2), \quad (93)$$

and the filter frequency characteristics is given by

$$H(\omega_1, \omega_2) = \frac{P_{f_N}(\omega_1, \omega_2)}{P_w(\omega_1, \omega_2)}, \qquad (94)$$

where $P(\omega_1, \omega_2)$ is the power spectral density of image. The filter in Equation (94) is called the noncausal Wiener filter.

Suppose that $f(n_1, n_2)$ is uncorrelated with $u(n_1, n_2)$ have both processes are zero mean value; we then have Equations 95-98.

$$R_{f_N}(n_1, n_2) = R_f(n_1, n_2) \qquad (95)$$
$$R_w(n_1, n_2) = R_f(n_1, n_2) + R_u(n_1, n_2) \qquad (96)$$

and

$$P_{f_N}(\omega_1, \omega_2) = P_f(\omega_1, \omega_2) \qquad (97)$$
$$P_w(\omega_1, \omega_2) = P_f(\omega_1, \omega_2) + P_u(\omega_1, \omega_2). \qquad (98)$$

Thus, the frequency response of the Wiener filter is given by

$$H(\omega_1, \omega_2) = \frac{P_f(\omega_1, \omega_2)}{P_f(\omega_1, \omega_2) + P_u(\omega_1, \omega_2)} \qquad (99)$$

If the $f(n_1, n_2)$ and $u(n_1, n_2)$ are samples of a Gaussian random field, Equation (88) becomes a minimum mean square error (MMSE) estimation problem, and the Wiener filter in Equation (99) becomes the optimal minimum mean square error estimator.

Since the power spectra $P_f(w_1, w_2)$, $P_u(w_1, w_2)$, and $H(\omega_1, \omega_2)$ are all real and nonnegative, the Wiener filter would affect only the spectral magnitude but not the phase. If $P_u(\omega_1, \omega_2)$ approaches 0, $H(\omega_1, \omega_2)$ will approach 1. This indicates that the filter tends to preserve the high SNR frequency components. If $P_u(\omega_1, \omega_2)$ *approaches* infinity, $H(\omega_1, \omega_2)$ will approach 0. The filter tends to attenuate the low SNR frequency components.

The Wiener solution is applied at each pixel in the image using only information from a surrounding neighborhood of pixels [97], [98]. In general image processing literature, this approach is often called the local minimum mean square error (MMSE). To circumvent the difficulties and limitations of the Wiener filtering, many image processing algorithms rely on local statistics (that is, estimates computed at each point using only data from a small surrounding region). Nagao and Matsuyama [99] proposed an iterative scheme that computes the sample mean and variance for nine differently oriented "mask" regions around the pixel. The pixel value at each point is then replaced by the sample mean from the region with the smallest variance. This entire process is repeated until there is no change of pixel values. This technique works well for images with sharp, high-amplitude edges, but not well for small or subtle structures. In addition, the shape of the mask regions employed can cause geometric distortion.

Another approach has been proposed by Song et al. in [100]. Each pixel is replaced by a function of the surrounding pixel, but the size of the surrounding region varies according to the estimate of the "signal activity" in that region. For flat and homogeneous regions, a large "window" is used, while for regions with edges a small window is used.

In order to overcome the difficulties and limitations of the Wiener filtering, the Auto Regressive (AR)-based interpolators were proposed to be used as an estimator of the noise variance in image [101], [102]. The result is a novel version of Wiener, called AR-Wiener filtering.

### B. AR-WIENER FILTER

Assume that there are 2N-point autocorrelation sequences $r(-N)$, $r(-N+1)$, ... $r(-1)$, $r(1)$, $r(2)$, ..., $r(N)$ within it the zero lag autocorrelation sample $r(0)$, represents the autocorrelation sample at the zero offset. The objective is to estimate the missing sample $r(0)$, using the remaining $2N$ samples and an AR model of the signal. The AR-Wiener Filter includes a least square error interpolation based on an autoregressive model. After obtaining the estimate of noise variance of the noisy image, it is then used as input parameter for the Wiener filter [103].

In order to implement this AR-Wiener Filter, a specialized SEM platform was used, as shown in Fig. 28.

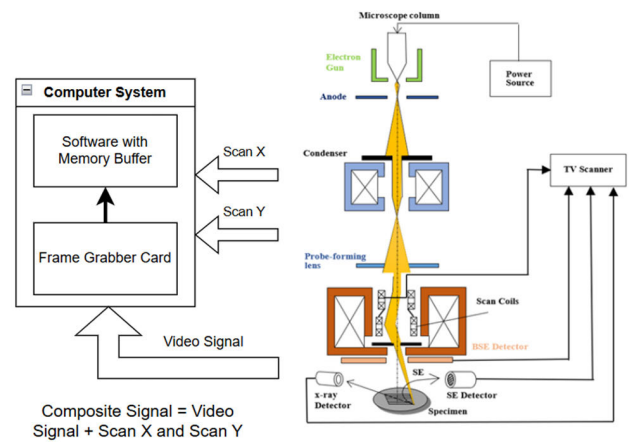

**FIGURE 28.** SEM experimental setup for real-time noise reduction system.

### C. COMPARATIVE ANALYSIS OF SNR ENHANCEMENT TECHNIQUES

Different SNR enhancement techniques offer varying levels of performance depending on the noise characteristics, image complexity, and computational requirements. Table 7 compares the key methods discussed in this review.

As seen in the table, machine learning-based approaches, particularly CNNs, provide the highest SNR improvement with minimal loss of detail. However, these methods are computationally intensive and require large datasets for training. On the other hand, traditional filtering methods like Gaussian and median filters are less effective at preserving details but are computationally simpler and widely applicable.

## X. FUTURE DIRECTIONS AND CHALLENGES

Despite significant advancements in SNR enhancement techniques for SEM, several challenges remain. One of the key issues is the trade-off between noise reduction and detail preservation. Many filtering techniques, particularly linear methods, tend to blur fine features along with the noise. Additionally, most noise reduction methods assume stationary noise, which may not always be the case in SEM.

**TABLE 7.** Comparison among SNR enhancement techniques.

| Technique | SNR Improvement | Detail Preservation | Computational Complexity | Applicability |
|---|---|---|---|---|
| Gaussian Filtering | Moderate | Low | Low | General Purpose |
| Median Filtering | High | Moderate | Moderate | Salt-and-Pepper Noise |
| Wavelet Denoising | High | High | High | High-Resolution Images |
| CNN-Based Denoising | Very High | Very High | Very High | Complex Noise Patterns |

Future research should focus on developing adaptive and hybrid approaches that can dynamically adjust to varying noise characteristics across different regions of the SEM image. The integration of machine learning with traditional methods offers a promising direction for achieving better SNR performance.

## XI. CONCLUSION

Signal-to-noise ratio is a critical factor that determines the quality and utility of SEM images. This survey has reviewed the key sources of noise in SEM, methods for estimating SNR, and various noise reduction techniques. While traditional methods such as linear and non-linear filtering remain relevant, modern approaches like machine learning-based denoising provide significant improvements in SNR while preserving fine details.

By understanding the strengths and limitations of each technique, researchers and practitioners can make informed decisions when selecting the appropriate SNR enhancement methods for their specific applications.

## ACKNOWLEDGMENT

The authors would like to express their sincere gratitude to Multimedia University for providing the necessary resources and a research environment to facilitate this work.

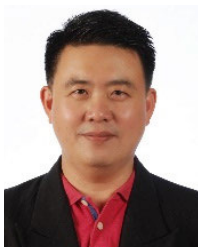

**KOK SWEE SIM** (Senior Member, IEEE) is currently a Professor with Multimedia University, Malaysia. He actively collaborates with various local and international universities and hospitals. He has filed 23 patents and 85 software copyrights. He is a fellow of the Academy of Sciences Malaysia, the Institution of Engineers Malaysia (IEM), and the Institution of Engineering and Technology (IET), U.K. Over the years, he has received numerous prestigious national and international awards. These include Japan Society for the Promotion of Science (JSPS) Fellowship, in 2018, the Top Research Scientists Malaysia (TRSM) Award from the Academy of Sciences Malaysia, in 2014, and Korean Innovation and Special Awards, in 2013, 2014, and 2015. He was a recipient of the TM Kristal Award and multiple World Summit on the Information Society (WSIS) Prizes, in 2017, 2018, 2019, 2020, and 2021.

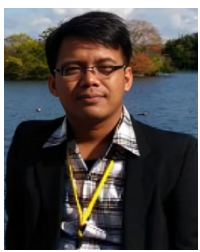

**IKSAN BUKHORI** (Member, IEEE) received the bachelor's degree in control system from President University, Indonesia, in 2013, and the Master of Philosophy degree in electronic system engineering from Universiti Teknologi Malaysia, Malaysia, in 2017. He is currently a Lecturer with the Electrical Engineering Department, Faculty of Engineering, President University, Indonesia. His research interests include control systems, robotics, artificial intelligence, and embedded systems.

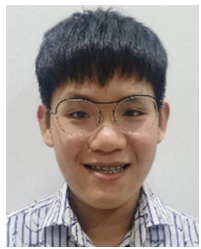

**DOMINIC CHEE YONG ONG** (Student Member, IEEE) received the Bachelor of Engineering degree (Hons.) in electronics, majoring in robotics and automation from Multimedia University, Malaysia, in 2024. He is currently pursuing the Master of Applied Engineering degree in IoT systems and technologies with Monash University, Malaysia.

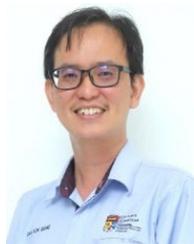

**KOK BENG GAN** (Member, IEEE) received the Bachelor of Science degree (Hons.) in material physics from Universiti Technologi Malaysia, in 2001, and the Ph.D. degree in electrical, electronic and system engineering from Universiti Kebangsaan Malaysia, in 2009. He was an Engineer in the field of electronic manufacturing services and original design manufacturing before venturing into academic research in 2005. He specializes in embedded system and artificial intelligent in healthcare. He is currently an Associate Professor with the Department of Electrical, Electronic and Systems Engineering, Faculty of Engineering and Built Environment, Universiti Kebangsaan Malaysia. His current research interests include biomedical optics and optical instrumentation, embedded system and signal processing for medical application and biomechanics and human motion analysis.

• • •